\documentclass[10pt,twocolumn,letterpaper]{article}

\usepackage{cvpr}              
\definecolor{cvprblue}{rgb}{0.21,0.49,0.74}
\usepackage[pagebackref,breaklinks,colorlinks,allcolors=cvprblue]{hyperref}
\usepackage{amsmath,amssymb,amsfonts}
\usepackage{algorithm}
\usepackage{algpseudocode}

\usepackage{graphicx}
\usepackage{textcomp}
\usepackage{xcolor}
\usepackage{multirow}
\usepackage{booktabs}
\usepackage{makecell}
\usepackage{subcaption}
\usepackage{pifont}

\newtheorem{definition}{Definition}

\def\confName{CVPR}
\def\confYear{2026}

\title{BD-Merging: Bias-Aware Dynamic Model Merging with\\ Evidence-Guided Contrastive Learning}

\author{
     Yuhan Xie\textsuperscript{\rm 1},
	Chen Lyu\textsuperscript{\rm 1}\thanks{Corresponding author.}\\
	\textsuperscript{\rm 1} MoE Key Laboratory of Interdisciplinary Research of Computation and Economics,\\ Shanghai University of Finance and Economics\\
	{\tt\small yhtse@stu.sufe.edu.cn, lyu.chen@sufe.edu.cn}
}

\begin{document}
\maketitle
\begin{abstract}

Model Merging (MM) has emerged as a scalable paradigm for multi-task learning (MTL), enabling multiple task-specific models to be integrated without revisiting the original training data. Despite recent progress, the reliability of MM under test-time distribution shift remains insufficiently understood. Most existing MM methods typically assume that test data are clean and distributionally aligned with both the training and auxiliary sources. However, this assumption rarely holds in practice, often resulting in biased predictions with degraded generalization.  To address this issue, we present BD-Merging, a bias-aware unsupervised model merging framework that explicitly models uncertainty to achieve adaptive reliability under distribution shift. First, BD-Merging introduces a joint evidential head that learns uncertainty over a unified label space, capturing cross-task semantic dependencies in MM. Second, building upon this evidential foundation, we propose an Adjacency Discrepancy Score (ADS) that quantifies evidential alignment among neighboring samples. Third, guided by ADS, a discrepancy-aware contrastive learning mechanism refines the merged representation by aligning consistent samples and separating conflicting ones. Combined with general unsupervised learning, this process trains a debiased router that adaptively allocates task-specific or layer-specific weights on a per-sample basis, effectively mitigating the adverse effects of distribution shift.   Extensive experiments across diverse tasks demonstrate that BD-Merging achieves superior effectiveness and robustness compared to state-of-the-art MM baselines.

\end{abstract}     
\section{Introduction}

Multi-task learning (MTL) provides an efficient paradigm for reducing storage and deployment costs by enabling a single model to perform multiple tasks through parameter sharing~\cite{kollias2024distribution,agiza2024mtlora}.
Despite its success, deploying traditional MTL in real-world systems remains challenging.
Training a unified network across heterogeneous tasks often requires substantial high-quality data and heavy computational resources~\cite{fifty2021efficiently,albalak2024improving}.
Moreover, MTL assumes concurrent access to all task-specific datasets, which is often infeasible due to proprietary or privacy constraints~\cite{yang2023adamerging}.
Such limitations have motivated growing interest in post-training integration, which combines independently trained task models without costly retraining.
Leveraging the proliferation of pre-trained and fine-tuned checkpoints~\cite{wolf2019huggingface,yang2024model}, \textbf{model merging (MM)} has emerged as an effective means of consolidating knowledge from multiple models into a single unified network.

While MM offers a scalable alternative to traditional MTL, most existing approaches rely on the strong assumption that the test-time data distribution remains consistent with both task-specific training domains and the auxiliary sources used during MM~\cite{yang2024representation,du2024parameter,lu2024twin,akiba2025evolutionary}.
In practice, such an assumption is rarely valid. Real-world test data are often subject to distribution shift, which disrupts the representational alignment among merged models and consequently restricts the adaptability and robustness of MM.
Such distribution shift typically manifests in two major forms: (1) \emph{test-time bias,} reflecting intra-task corruption and domain-specific heterogeneity, and (2) \emph{generalization to unseen tasks}, reflecting inter-task distribution discrepancy.

\begin{figure*}[!t]
	\centering
	\begin{subfigure}[t]{0.48\textwidth}
		\centering
		\includegraphics[width=\textwidth]{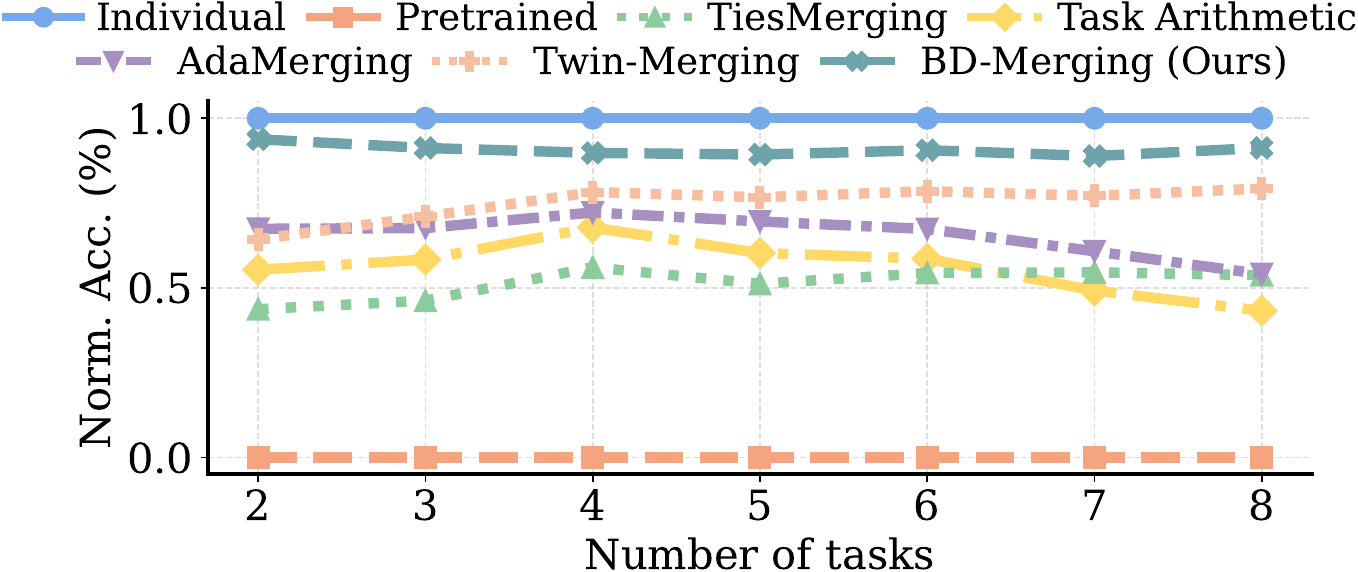}
		\caption{Across varying numbers of tasks on data with test-time bias.}
		\label{fig:task_number}
	\end{subfigure}
	\hspace{3mm}
	\begin{subfigure}[t]{0.48\textwidth}
		\centering
		\includegraphics[width=\textwidth]{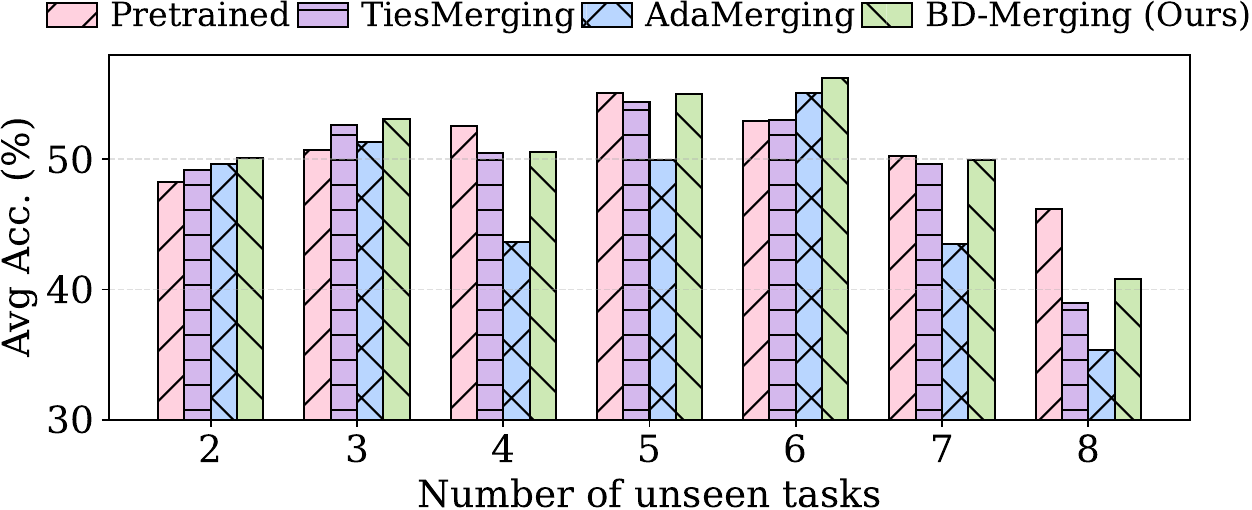}
		\caption{Across varying unseen task numbers on clean data.}
		\label{fig:task_unseen}
	\end{subfigure}
	\vspace{-2mm}
	\caption{Averaged normalized accuracy under different distribution shift settings across multiple benchmarks.}
	\label{fig:task_comparison}
\end{figure*}

The first form, \textbf{test-time bias}, arises from natural corruption (e.g., sensor noise and transmission distortions) and domain-specific heterogeneity caused by environmental variations~\cite{hendrycks2019benchmarking,mintun2021interaction}. As shown in Figure~\ref{fig:task_number}, such corruption shifts test-time inputs away from the merging distributions.
Consequently, the robustness of MM strategies deteriorates, leading to reduced overall performance~\cite{tian2023modeling}.

The second form concerns \textbf{generalization to unseen tasks}, which reflects an inter-task discrepancy that arises when the merged model encounters tasks or domains not represented during merging.
As illustrated in Figure~\ref{fig:task_unseen}, several MM strategies (e.g., AdaMerging) exhibit substantial performance degradation on unseen tasks compared with their respective pretrained models, thereby revealing a critical reliability risk for real-world deployment.
To improve merging quality, recent works have attempted to leverage auxiliary data or external guidance signals to align model parameters more effectively~\cite{tang2023concrete,lu2024twin}.
However, when the distribution of auxiliary data diverges from that of the target tasks, such reliance can further amplify the distributional gap ~\cite{wei2025modeling}. This, in turn, results in overfitting to seen domains and deteriorated generalization to previously unseen tasks.

These phenomena highlight two fundamental challenges of MM under distribution shift:
(1) \textbf{Conflicting knowledge and biased integration}, where test-time bias leads to inconsistent model behaviors that undermine the robustness of the merged model.
(2) \textbf{Limited cross-task generalization}, which restricts adaptation to unseen tasks. Existing methods struggle to capture the fine-grained, sample-level discrepancies required for reliable alignment under such distribution shifts,  resulting in limited robustness and generalization for MM~\cite{yadav2023resolving,yang2024representation,yang2023adamerging,davari2024model}.

\begin{figure*}[!t]
	\centering
	\includegraphics[width=1\textwidth]{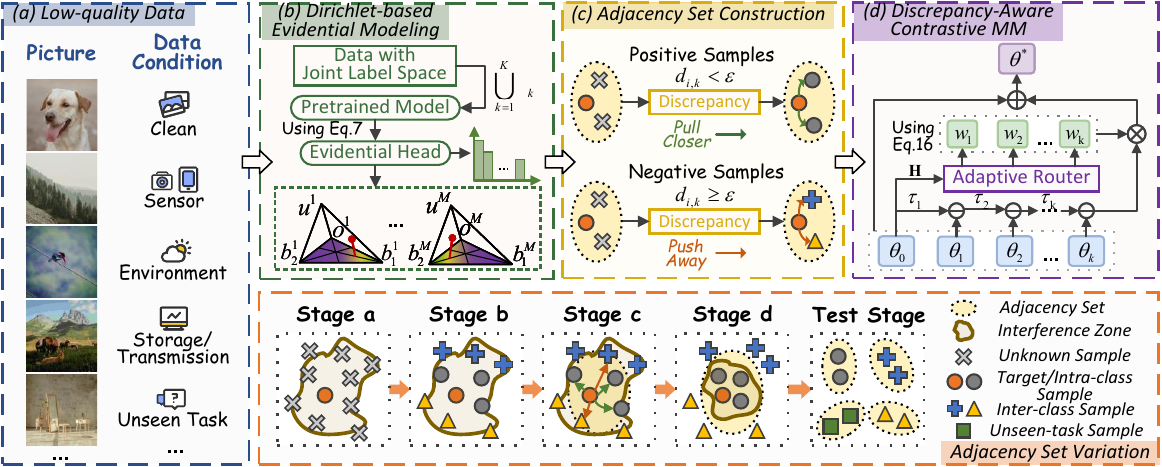}
	\vspace{-5mm}
	\caption{The process of BD-Merging. 
		BD-Merging includes three modules: (1) Modeling uncertainty using a Dirichlet-based evidential head; (2) Constructing an adjacency set and calculating the ADS to measure evidential alignment; and (3) Performing discrepancy-aware contrastive MM with a debiased router, guided by ADS. Stage (a)-(d) demonstrates the evolution of samples in the feature space through various stages.	}
	\vspace{-3mm}
	\label{fig:BDMerging}
\end{figure*}

To address these challenges, we propose \textbf{BD-Merging}, a bias-aware and adaptive MM framework that dynamically adjusts model behavior at test time to improve robustness and generalization, as illustrated in Figure~\ref{fig:BDMerging}.
The core insight is to exploit evidential uncertainty to capture distributional discrepancies and guide adaptive representation alignment during MM.
Specifically, BD-Merging incorporates a \textbf{joint evidential head} into a pretrained backbone to perform uncertainty modeling via 
\emph{Evidential Deep Learning (EDL)} \cite{gao2025comprehensive}. This head produces fine-grained evidence that represents category-wise support for each prediction, enabling the merged model to capture cross-task semantic dependencies while identifying uncertainty signals that indicate potential distribution shift.
Based on these evidential outputs, we define an \textbf{Adjacency Discrepancy Score (ADS)} to quantify alignment among neighboring samples within a local adjacency set. Guided by ADS, BD-Merging constructs positive and negative sample pairs, and applies a \textbf{discrepancy-aware contrastive learning} strategy that strengthens the consistency of reliable samples while separating conflicting ones.
This process effectively enhances the model's ability to distinguish in-distribution data from corrupted or unseen task inputs.
Finally, BD-Merging trains a debiased router by integrating discrepancy-aware contrastive learning with unsupervised optimization. The router adaptively allocates task-specific or layer-specific weights on a per-sample basis, thereby dynamically constructing shared knowledge based on test-time inputs to mitigate distribution shifts. Our main contributions are summarized as follows:
\begin{itemize}
\item We revisit the reliability of MM under test-time distribution shift and identify two key challenges: conflicting knowledge and biased integration, and limited cross-task generalization. 
\item  We propose BD-Merging, a bias-aware MM framework that explicitly models sample-level bias via evidential uncertainty. It introduces a joint evidential head for uncertainty modeling, an ADS for evidential alignment, and a discrepancy-aware contrastive learning that guides a debiased router for adaptive weight allocation.
\item Extensive experiments demonstrate that BD-Merging achieves superior robustness and generalization compared to state-of-the-art MM baselines, approaching the performance of individually fine-tuned models while maintaining efficiency for real-world deployment.
\end{itemize}

\section{Related Work} \label{Related_Work}

\subsection{Model Merging}
MM integrates independently fine-tuned task-specific models into a unified multi-task model without accessing raw data or requiring retraining \cite{yang2024model}. Existing approaches primarily focus on mitigating interference among models and handling data heterogeneity during inference \cite{lu2024twin}. FisherMerging \cite{matena2022merging} and RegMean \cite{jin2022dataless} rely on Fisher information and inner-product matrices for weighted merging, but require auxiliary statistics that are costly to obtain. Task Arithmetic \cite{ilharco2022editing} extends simple weight averaging to more general arithmetic operations in parameter space, enabling finer control over merged model behavior. To further alleviate task interference arising from incompatible parameter updates, several advanced methods have been proposed. Among them, Ties-Merging \cite{yadav2023ties} mitigates parameter interference by resolving redundancy and sign conflicts, AdaMerging \cite{yang2023adamerging} learns task-adaptive merging coefficients, and DARE \cite{yu2024language} promotes sparse fusion for reduced interference and improved generalization.

Despite recent advances, most MM methods face two key challenges related to test-time distribution shifts: test-time bias and generalization to unseen tasks. However, existing methods largely neglect these distribution shifts, resulting in limited scope and effectiveness for MM. These challenges highlight the need for merging strategies that explicitly address test-time distribution shifts.



\subsection{Evidential Deep Learning}
EDL \cite{sensoy2018evidential} enables fine-grained uncertainty quantification for Deep Neural Networks (DNNs) by modeling the underlying evidence behind predictions \cite{chen2022evidential, yu2024evidential}. It is particularly effective in handling distribution shift, where the total evidence provides a summation-based, non-competitive confidence measure \cite{zheng2023evidential}. Recent formulations interpret network outputs as parameters of a Dirichlet distribution \cite{deng2023uncertainty}, forming the basis of evidential neural networks.

The Dirichlet-based EDL framework has been successfully extended to various domains, including multi-view learning \cite{han2022trusted,xu2024reliable}, unsupervised domain adaptation \cite{chen2022evidential, pei2024evidential}, out-of-distribution detection \cite{qu2024hyper,aguilar2023continual}, regression \cite{ye2024uncertainty,wu2024evidence}, and action understanding \cite{bao2021evidential,gao2023vectorized}. 
In this work, we extend Dirichlet-based EDL to address real-world test-time distribution shifts, aiming to provide more reliable guidance for the MM process.


\section{Preliminaries}\label{Preliminaries}
\subsection{Problem Setup}

Let \( K \) denote the number of supervised fine-tuned task models \( \{\theta_k\}_{k=1}^K \) involved in the MM process, where each \( \theta_k \) is fine-tuned from a shared pretrained model \( \theta_0 \). We define the corresponding task vector as \( \tau_k = \theta_k - \theta_0 \). Each task \( T_k \) is associated with a label space \( \mathcal{Y}_k = \{1, 2, \ldots, C_k\} \), where \( C_k \) denotes the number of classes.

We assume access to an unannotated auxiliary dataset $D^{\text{aux}} = \{(x_i)\}_{i=1}^N$.  $D^{\text{aux}}$ is used to learn a set of task-adaptive weights \( \{w_k\}_{k=1}^K \) or layer-adaptive weights $
\{ \{ w_k^{(l)} \}_{l=1}^{L} \}_{k=1}^{K}$ through a  training procedure. \( \{w_k\}_{k=1}^K \) are then used to merge supervised fine-tuned models into a unified model \( f_{\theta^*} \) via a merging method \( \mathcal{M} \), defined as $ \theta^* = \mathcal{M}(\theta_0,\{\tau_k\}_{k=1}^K, \{w_k\}_{k=1}^K)$.

Let \( \mathcal{T}^{\text{te}} \) denote the set of evaluation tasks, where each task \( T_t \in \mathcal{T}^{\text{te}} \) has a test set \( D_t^{\text{te}} = \{(\tilde{x}_j^t, y_j^t)\}_{j=1}^{n_t} \), potentially containing input-label pairs under distribution shifts. The goal of BD-Merging is to optimize 
merging weights such that the average test loss across evaluation tasks is minimized:
\[
\min_{W^*} \quad \frac{1}{|\mathcal{T}^{\text{te}}|} \sum_{T_t \in \mathcal{T}^{\text{te}}} \frac{1}{n_t} \sum_{j=1}^{n_t} \ell(f_{\theta^*}(\tilde{x}_j^t), y_j^t),
\]
where \( \ell(\cdot, \cdot) \) denotes a task-specific loss function.


\section{Methodology}

\subsection{Dirichlet-Based Evidential Modeling}

The Dirichlet-based EDL framework~\cite{sensoy2018evidential} represents class probabilities using Dirichlet distributions. The concentration parameters \( \boldsymbol{\alpha} \) quantify class-specific evidence, enabling uncertainty-aware predictions grounded in Subjective Logic~\cite{jsang2018subjective} and Dempster-Shafer Theory~\cite{dempster1968generalization,shafer1976mathematical}.

\begin{definition}[Dirichlet Distribution]
	\label{def:dirichlet}
	The Dirichlet distribution is defined over the probability simplex $	\Delta^{C_{k}-1}_p = \left\{ \mathbf{p} \in \mathbb{R}^{C_{k}} \;\middle|\; \sum_{c=1}^{C_{k}} p_c = 1,\, 0 \leq p_c \leq 1 \right\},$
	and is parameterized by \( \boldsymbol{\alpha} = [\alpha_1, \dots, \alpha_{C_{k}}] \) with \( \alpha_c > 0 \). The probability density is given by:	
	\begin{equation}
		\mathrm{Dir}(\mathbf{p} \mid \boldsymbol{\alpha}) =
		\begin{cases}
			\frac{1}{B(\boldsymbol{\alpha})} \prod\limits_{c=1}^{C_k} p_c^{\alpha_c - 1}, & \text{if } \mathbf{p} \in \Delta^{C_{k}-1}_p, \\
			0, & \text{otherwise},
		\end{cases}
	\end{equation}
	where \( B(\boldsymbol{\alpha}) \) is the multivariate beta function.
\end{definition}

\begin{definition}[Dirichlet-Based Evidential Deep Learning]
	\label{def:dirichlet-edl}
	In EDL, the  outputs of DNNs are interpreted as class-specific evidence \( \mathbf{e} = (e_1, \dots, e_{C_k})^\top \), where \( e_c \geq 0 \). The corresponding Dirichlet parameters are defined as \( \boldsymbol{\alpha} = \mathbf{e} + \mathbf{1} \), leading to a Dirichlet strength \( S = \|\mathbf{e}\|_1 + C_k \).
	
	Following Subjective Logic, the belief mass \( b_c \), uncertainty \( u \), and final predictive probability \( p_c \) for class \( c \) are defined as:
	\begin{equation}
		b_c = \frac{e_c}{S}, \quad 
		u = \frac{C_k}{S}, \quad 
		p_c = b_c + u \cdot \frac{1}{C_k} = \frac{\alpha_c}{S}.
	\end{equation}
\end{definition}

\subsection{Joint Evidential Head}

Test-time data in MM are often subject to distribution shifts, amplifying semantic ambiguities due to overlapping label spaces and inter-task dependencies, resulting in inconsistent predictions and reduced model reliability. 
To address these issues, we propose a joint evidential head integrated into a pretrained backbone $\theta_0$, operating over a unified label set \(\mathcal{Y} = \bigcup_{k=1}^K \mathcal{Y}_k\), where \(\mathcal{Y}_k\) denotes the label space of the \(k\)-th task and \(|\mathcal{Y}| = L\).

Although structural unification provides a shared representation space, conventional uncertainty metrics in EDL are limited in capturing semantic shift under cross-task conditions. Total evidence reflects prediction strength but neglects inter-class relationships~\cite{sensoy2018evidential,chen2022evidential}, while predicted-class confidence focuses only on the most likely class, ignoring class competition and ambiguity~\cite{pei2024evidential}.

To address this, we introduce inter-class evidential contrast (IEC) to integrate inter-class semantic dependencies and class competition for enhanced uncertainty estimation:
\begin{equation}
	\nu = \left( \frac{S}{\alpha_{\hat{c}^{(1)}}} \right) \cdot \left( \frac{L}{S} \right) \cdot  \left( \frac{\alpha_{\hat{c}^{(2)}}}{\alpha_{\hat{c}^{(1)}}} \right),
\end{equation}
where \( \hat{c}^{(i)} \) denotes the \( i \)-th most probable class.

To enforce an inverse relationship between the uncertainty $u$ and IEC, we leverage an inverse correlation loss that penalizes their inconsistency, encouraging high IEC values when $u$ is low and suppressing IEC when $u$ is high:
\begin{equation}
	\mathcal{L}_{\mathrm{Inv}} = - \sum_{i=1}^N \left[ \nu_i \log(1 - u_i) + (1 - \nu_i) \log u_i \right].
\end{equation}

Furthermore, the joint evidential head is trained with an entropy-based unsupervised objective, regularized by a KL divergence to a non-informative prior:
\begin{equation}
	\mathcal{L}_{\mathrm{Ent}} = \sum_{i=1}^N \left[ \sum_{j=1}^L p_{ij} \log p_{ij} + \lambda \cdot \mathrm{KL}\big( \mathrm{Dir}(\boldsymbol{\alpha}_i) \,\|\, \mathrm{Dir}(\mathbf{1}) \big) \right],
\end{equation}
where \( \boldsymbol{\alpha}_i = (\alpha_{i1}, \ldots, \alpha_{iL}) \) are the Dirichlet concentration parameters, \( p_{ij} = \alpha_{ij} / S_i \) denotes the predictive probability. The coefficient \( \lambda \) controls the regularization strength.

To jointly optimize predictive accuracy and epistemic uncertainty calibration, we combine the entropy-based unsupervised loss with the inverse correlation loss:
\begin{equation} \label{eq6}
	\mathcal{L}_{\mathrm{Head}} = \mathcal{L}_{\mathrm{Ent}} + \gamma \mathcal{L}_{\mathrm{Inv}},
\end{equation}
where \( \gamma \) controls the trade-off between the loss terms.

\subsection{Adjacency Set Construction}
Based on outputs from the joint evidential head, we introduce the Adjacency Discrepancy Score (ADS) to quantify local evidential alignment among neighborhood samples. For each sample \( x_i \), we define its adjacency set as \( \mathcal{A}_r(x_i) = \{x_i\} \cup \mathcal{N}_r(x_i) \), where \( \mathcal{N}_r(x_i) \) denotes the radius-\( r \) neighborhood in the feature space.

ADS captures localized evidential discrepancies by jointly evaluating three complementary factors centered on a target sample \( x_i \): (i) prediction sharpness, reflecting the overall epistemic uncertainty within its neighborhood; (ii) semantic divergence, quantifying class-level inconsistency between \( x_i \) and 
its adjacent samples; and (iii) opinion conflicts, measuring belief disagreement between \( x_i \) and a specific neighbor \( x_k \). Together, these components provide a unified view of local uncertainty dynamics.
\vspace{-1mm}
\begin{algorithm}[t]
	\caption{BDMerging}
	\label{alg:BDMerging}
	\begin{algorithmic}[1]
	\Statex \textbf{Input:} Backbone $\theta_0$, task vectors $\{\tau_k\}_{k=1}^K$, unified label set $\mathcal{Y}$, 
	temperature $\tau$, radius $r$, threshold $\epsilon$, coefficients $(\lambda, \gamma, \eta)$
	\Statex \textbf{Output:} Merged parameters $\theta^*$
	\State \textcolor{blue}{\textbf{Step 1: Joint Evidential Head}}
	\For{each sample $x_i$}
  	\State Compute evidential statistics and derive $\nu$
	\State Accumulate contributions to $\mathcal{L}_{\mathrm{Inv}}$ and $\mathcal{L}_{\mathrm{Ent}}$
	\EndFor
	\State Optimize the prediction head to minimize $\mathcal{L}_{\mathrm{Head}}$
	\State \textcolor{blue}{\textbf{Step 2: Adjacency Set Construction}}
	\For{each sample $x_i$}
	\State Identify the adjacency set \( \mathcal{A}_r(x_i) = \{x_i\} \cup \mathcal{N}_r(x_i) \)
	\For{each neighbor $x_k \in \mathcal{N}_r(x_i)$}
	\State Compute ADS $d_{ik}$ using complementary factors: $\mathrm{Sharp}(x_i)$, $\mathrm{Div}(x_i)$, $\mathrm{Conf}(x_i, x_k)$
	\EndFor
	\EndFor
	\State \textcolor{blue}{\textbf{Step 3: Discrepancy-Aware Contrastive Merging}}
	\For{each anchor $x_i$}
	\State Partition $\mathcal{N}_r(x_i)$ into $\mathcal{M}_r^+(i)$ and $\mathcal{M}_r^-(i)$ by $d_{ik}$
	\EndFor
	\State Compute $\mathcal{L}_{\mathrm{BD}}$ to optimize router weights $w_k$	
	\State \Return Merged parameters $\theta^*=\theta_0+\sum_k w_k\cdot\tau_k$
	\end{algorithmic}
\end{algorithm}

\paragraph{Prediction Sharpness.} 
This term evaluates the concentration of evidence within the \( \mathcal{A}_r(x_i) \), indicating the overall epistemic strength of neighboring predictions:
\begin{equation} \label{eq7}
	\mathrm{Sharp}(x_i) = \mathbb{E}_{x_j \in \mathcal{A}_r(x_i)} \log \left( \frac{S_j}{\max_c \alpha_{jc}} - 1 \right),
\end{equation}

\paragraph{Semantic Divergence.}
This term quantifies the class-level distributional deviation between the target sample and its neighbors, capturing cross-sample semantic inconsistency:
\begin{equation} \label{eq8}
	\mathrm{Div}(x_i) = \mathbb{E}_{x_j \in \mathcal{N}_r(x_i)} \left\| \frac{\boldsymbol{\alpha}_i}{S_i} - \frac{\boldsymbol{\alpha}_j}{S_j} \right\|_1,
\end{equation}

\paragraph{Opinion Conflicts.}
This term quantifies the belief-level conflicts between \( x_i \) and a specific neighbor \( x_k \), weighted by mutual confidence levels:
\begin{equation} \label{eq9}
	\mathrm{Conf}(x_i,x_k) =  \sum_{c=1}^{L} \left| p_{ic} - p_{kc} \right| \cdot (1 - u_i)(1 - u_k),
\end{equation}

We denote the ADS between a target sample \( x_i \) and its neighbor \( x_k \in \mathcal{N}_r(x_i) \) by \( d_{ik} \), which is computed as the product of three components:
\begin{equation}
	d_{ik} = \mathrm{Sharp}(x_i) \cdot \mathrm{Div}(x_i) \cdot \mathrm{Conf}(x_i, x_k).
\end{equation}

\subsection{Discrepancy-Aware Contrastive Merging}  

\paragraph{Debiased Router.}
A model merged with a single set of weights is susceptible to task interference \cite{gargiulo2025task}. To mitigate such interference among heterogeneous tasks and robustly integrate their contributions, we introduce a debiased router that dynamically assigns task- or layer-specific  weights, guiding the MM process in a data-driven manner.

Let $\mathbf{H} = \{h_i\}_{i=1}^N$ denote the set of last-layer token embeddings obtained from $\theta_0$, where each $h_i$ represents the sequence of token embeddings for sample $x_i$. The router computes a  weight vector based on $\mathbf{H}$:
\begin{equation}
	\{w_k\}_{k=1}^K = \text{softmax}(R(\mathbf{H})),
\end{equation}
Where $R$ denotes the router network. The merged model parameters are given by $ \theta^* = \theta_0 + \sum_{k=1}^K w_k \cdot \tau_k$.

\paragraph{Discrepancy-aware Contrastive Loss.}
To enhance representation learning under distribution shift conditions,  we propose a discrepancy-aware contrastive loss that leverages evidential discrepancies to adaptively guide semantic alignment during MM.  Unlike traditional contrastive objectives, our approach dynamically partitions each sample's adjacency set based on evidential levels, thereby improving robustness to test-time distribution shift.

\begin{table*}[t]
	\centering
	\caption{Performance comparison of MM methods under varying intensities of test-time bias. \textbf{bold} and \textcolor{blue}{\underline{Blue}}  numbers indicate the best and second-best results, respectively.}
	\vspace{-1mm}
	\resizebox{\textwidth}{!}{
		\begin{tabular}{lll|cccccccc|c}
			\toprule
			\textbf{Method} & \textbf{Type} & \textbf{Level} 
			& \textbf{SUN397 \cite{xiao2016sun}} & \textbf{Cars \cite{krause20133d}} & \textbf{RESISC45 \cite{cheng2017remote}} & \textbf{EuroSAT \cite{helber2019eurosat}} & \textbf{SVHN \cite{netzer2011reading}} & \textbf{GTSRB \cite{stallkamp2011german}} & \textbf{MNIST \cite{lecun2010mnist}} & \textbf{DTD \cite{cimpoi2014describing}} & \textbf{Avg Acc.} \\
			\midrule
			\multirow{4}{*}{\shortstack{Task\\Arithmetic \cite{ilharco2022editing}}} 
			& Clean & -& 55.37 &54.98 & 66.10& 78.40 & 80.21 &69.65 &97.34  &50.64 & 69.09 \\		
			& \multirow{3}{*}{Corrupted} 
			& $L_1$ &53.07\scriptsize($\downarrow$4.2) &52.68\scriptsize($\downarrow$4.2)&63.90\scriptsize($\downarrow$3.3) &67.15\scriptsize($\downarrow$14.3) & 71.33\scriptsize($\downarrow$11.1) &64.09\scriptsize($\downarrow$8.0) &92.04\scriptsize($\downarrow$5.4) &48.30\scriptsize($\downarrow$4.6) &64.07\scriptsize($\downarrow$7.3)    \\

			&& $L_2$ &50.57\scriptsize($\downarrow$8.7) 
			&49.24\scriptsize($\downarrow$10.4)
			&57.60\scriptsize($\downarrow$12.9) 
			&63.35\scriptsize($\downarrow$19.2) 
			&66.68\scriptsize($\downarrow$16.9) 
			&59.09\scriptsize($\downarrow$15.2) 
			&88.67\scriptsize($\downarrow$8.9) 
			&47.23\scriptsize($\downarrow$6.7) 
			&60.30\scriptsize($\downarrow$12.7)    \\
			
			&& $L_3$ &47.35\scriptsize($\downarrow$14.5) 
			&46.40\scriptsize($\downarrow$15.6)
			&53.50\scriptsize($\downarrow$19.1) 
			&62.05\scriptsize($\downarrow$20.9) 
			&64.81\scriptsize($\downarrow$19.2) 
			&56.75\scriptsize($\downarrow$18.5) 
			&85.71\scriptsize($\downarrow$11.9) 
			&43.40\scriptsize($\downarrow$14.3) 
			&57.50\scriptsize($\downarrow$16.8)    \\
			\midrule
			
			\multirow{4}{*}{\shortstack{Ties-Merging \cite{yadav2023ties}}} 
			& Clean & -&  64.57&64.42 &74.60 & 76.65 &81.28  &69.38 &96.53  &55.96 &72.92  \\
			& \multirow{3}{*}{Corrupted} 
			& $L_1$ &\textcolor{blue}{\underline{62.75\scriptsize($\downarrow$2.8)}} & 62.03\scriptsize($\downarrow$3.7)& 72.00\scriptsize($\downarrow$3.5) & 67.35\scriptsize($\downarrow$12.1) & 72.71\scriptsize($\downarrow$10.5) & 63.71\scriptsize($\downarrow$8.2) 
			&91.37\scriptsize($\downarrow$5.3) & 53.19\scriptsize($\downarrow$4.9) &68.14\scriptsize($\downarrow$6.6)    \\

			&& $L_2$ &{\textbf{60.43\scriptsize($\downarrow$6.4)}} 
			&58.90\scriptsize($\downarrow$8.6)
			&65.40\scriptsize($\downarrow$12.3) &63.15\scriptsize($\downarrow$17.6) & 69.08\scriptsize($\downarrow$15.0) &58.88\scriptsize($\downarrow$15.1) &87.85\scriptsize($\downarrow$9.0) &51.06\scriptsize($\downarrow$8.8) &64.34\scriptsize($\downarrow$11.8)    \\

			&& $L_3$ &57.00 \scriptsize($\downarrow$11.7) &55.22\scriptsize($\downarrow$14.3)&61.10\scriptsize($\downarrow$18.1) &61.15 \scriptsize($\downarrow$20.2) &67.09\scriptsize($\downarrow$17.5) &56.59\scriptsize($\downarrow$18.4) &85.82\scriptsize($\downarrow$11.1) &48.09\scriptsize($\downarrow$14.1) & 61.51\scriptsize($\downarrow$15.6)    \\
			\midrule
			
			\multirow{4}{*}{\shortstack{AdaMerging \cite{yang2023adamerging}\\(Task-wise)}} 
			& Clean & -& 59.09 & 57.18& 67.10 & 86.15 & 80.76 & 70.70 & 95.67 & 50.64 & 70.91 \\
			& \multirow{3}{*}{Corrupted} 
			& $L_1$ & {\textbf{57.45\scriptsize($\downarrow$2.8)}} & 55.35\scriptsize($\downarrow$3.2)& 65.20\scriptsize($\downarrow$2.8) & 75.30\scriptsize($\downarrow$12.6) & 71.88\scriptsize($\downarrow$11.0) & 64.40\scriptsize($\downarrow$8.9) & 90.00\scriptsize($\downarrow$5.9) & 48.51\scriptsize($\downarrow$4.2) & 66.01\scriptsize($\downarrow$6.9)    \\
			
			& & $L_2$ & \textcolor{blue}{\underline{55.20\scriptsize($\downarrow$6.6)}} & 52.54\scriptsize($\downarrow$8.1) & 58.20\scriptsize($\downarrow$13.3) & 70.20\scriptsize($\downarrow$18.5) & 67.43\scriptsize($\downarrow$16.5) & 59.32\scriptsize($\downarrow$16.1) & 86.20\scriptsize($\downarrow$9.9) & 46.81\scriptsize($\downarrow$7.6) & 61.99\scriptsize($\downarrow$12.6) \\
			
			& & $L_3$ & 51.54\scriptsize($\downarrow$12.8) & 49.05\scriptsize($\downarrow$14.2) &54.50\scriptsize($\downarrow$18.8)  & 68.00\scriptsize($\downarrow$21.1) & 64.60\scriptsize($\downarrow$20.0) & 57.66\scriptsize($\downarrow$18.4) & 84.37\scriptsize($\downarrow$11.8)&
			42.98\scriptsize($\downarrow$15.1)  & 59.09\scriptsize($\downarrow$16.7) \\
			\midrule
			
			\multirow{4}{*}{\shortstack{AdaMerging \cite{yang2023adamerging}\\(Layer-wise)}} 
			& Clean & - & 59.76 & 60.70 & 72.70 & 90.05 & 84.12 & 76.68 & 97.63& 52.98 & 74.33\\
			& \multirow{3}{*}{Corrupted} 
			& $L_1$ & 57.84\scriptsize($\downarrow$3.2) & 58.62\scriptsize($\downarrow$3.4)  & \textcolor{blue}{\underline{70.90\scriptsize($\downarrow$2.5)}}  & 76.30\scriptsize($\downarrow$15.3)  & 74.83\scriptsize($\downarrow$11.0) &70.60\scriptsize($\downarrow$7.9)  & 92.24\scriptsize($\downarrow$5.5)  & 50.64\scriptsize($\downarrow$4.4) &69.00\scriptsize($\downarrow$7.2)\\

			& & $L_2$ & 54.89\scriptsize($\downarrow$8.1) & 54.46\scriptsize($\downarrow$10.3) & 63.90\scriptsize($\downarrow$12.1) & 71.50\scriptsize($\downarrow$20.6) & 68.45\scriptsize($\downarrow$18.6) & 63.10\scriptsize($\downarrow$17.7) & 90.41\scriptsize($\downarrow$7.4) & 48.94\scriptsize($\downarrow$7.6) & 64.46\scriptsize($\downarrow$13.3) \\
			
			& & $L_3$  & 51.15\scriptsize($\downarrow$14.4) &51.44\scriptsize($\downarrow$15.3)&
			59.00\scriptsize($\downarrow$18.8) &68.25\scriptsize($\downarrow$24.2) &65.69\scriptsize($\downarrow$21.9) & 60.17\scriptsize($\downarrow$21.5) &89.56\scriptsize($\downarrow$8.3) & 44.47\scriptsize($\downarrow$16.1) & 61.22\scriptsize($\downarrow$17.6) \\
			\midrule
			
			\multirow{4}{*}{\shortstack{Ties-Merging\\w/Surgery \cite{yang2024representation}}} 
			& Clean & - & 68.84& 63.44 & 82.00 & 92.30 & 83.04 & 82.60 & 97.20 & 64.68 & 79.26 \\
			& \multirow{3}{*}{Corrupted} 
			& $L_1$ & 66.63\scriptsize($\downarrow$3.2) & 60.68\scriptsize($\downarrow$4.4) & 79.10\scriptsize($\downarrow$3.5) & 80.45\scriptsize($\downarrow$12.8) &74.81\scriptsize($\downarrow$9.9)  & 74.91\scriptsize($\downarrow$9.3)&
			92.49\scriptsize($\downarrow$4.8)  &\textcolor{blue}{\underline{64.47\scriptsize($\downarrow$0.3)}} & 74.19\scriptsize($\downarrow$6.4) \\
			
			& & $L_2$ & 64.17\scriptsize($\downarrow$6.8) & 57.41\scriptsize($\downarrow$9.5) & 71.90\scriptsize($\downarrow$12.3) & 74.90\scriptsize($\downarrow$18.9) & 69.57\scriptsize($\downarrow$16.2)& 69.87\scriptsize($\downarrow$15.4) & 90.52\scriptsize($\downarrow$6.9) & \textcolor{blue}{\underline{61.28\scriptsize\scriptsize($\downarrow$5.3)}}& 69.95\scriptsize($\downarrow$11.7) \\
			
			& & $L_3$  &\textcolor{blue}{\underline{60.36\scriptsize($\downarrow$12.3)}}  &53.67\scriptsize($\downarrow$15.4)  & 65.60\scriptsize($\downarrow$20.0) & 73.35\scriptsize($\downarrow$20.5) & 67.74\scriptsize($\downarrow$18.4) & 65.42\scriptsize($\downarrow$20.8) &89.39\scriptsize($\downarrow$8.0) &57.87\scriptsize($\downarrow$10.5)  & 66.67\scriptsize($\downarrow$15.9)\\
			\midrule

			\multirow{4}{*}{\shortstack{AdaMerging\\w/Surgery \cite{yang2024representation}}} 
			& Clean & - &  70.79 & 68.01 & 87.10 & 96.35 & 90.32& 94.35 & 98.48 & 69.79 & 84.40 \\
			& \multirow{3}{*}{Corrupted} 
			& $L_1$ & 
			66.75\scriptsize($\downarrow$5.7) & 65.76\scriptsize($\downarrow$3.3) & 82.70\scriptsize($\downarrow$5.1) & 86.15\scriptsize($\downarrow$10.6) & 82.19\scriptsize($\downarrow$9.0) & \textcolor{blue}{\underline{88.50\scriptsize($\downarrow$6.2)}} & 93.53\scriptsize($\downarrow$5.0) & 66.60\scriptsize($\downarrow$4.6) &79.02\scriptsize($\downarrow$6.4) \\

			& & $L_2$ & 65.04\scriptsize($\downarrow$8.1)& 61.87\scriptsize($\downarrow$9.0) & 75.50\scriptsize($\downarrow$13.3) & 79.70\scriptsize($\downarrow$17.3) & 76.26\scriptsize($\downarrow$15.6) & 80.37\scriptsize($\downarrow$14.8) & 91.00\scriptsize($\downarrow$7.6) & 64.89\scriptsize($\downarrow$7.0) & 74.33\scriptsize($\downarrow$11.9) \\
			
			& & $L_3$  & 59.74\scriptsize($\downarrow$15.6) & 58.55\scriptsize($\downarrow$13.9) & 70.10\scriptsize($\downarrow$19.5) &77.25\scriptsize($\downarrow$19.8)  &74.20\scriptsize($\downarrow$17.8)  &77.73\scriptsize($\downarrow$17.6)  & 89.58\scriptsize($\downarrow$9.0)& 60.64\scriptsize($\downarrow$13.1) &70.97\scriptsize($\downarrow$15.9) \\
			\midrule
			
			\multirow{4}{*}{\shortstack{Twin-Merging \cite{lu2024twin}}} 
			& Clean & - & 68.80& 68.77 & 85.27 & 96.65 & 89.16& 95.20 & 98.11& 70.85 & 84.10 \\
			& \multirow{3}{*}{Corrupted} 
			& $L_1$ 
			& 66.63\scriptsize($\downarrow$3.2) & 66.32\scriptsize($\downarrow$3.6) & 82.07\scriptsize($\downarrow$3.7) & 87.25\scriptsize($\downarrow$9.7) & 81.08\scriptsize($\downarrow$9.1) &88.39\scriptsize($\downarrow$7.2)  &93.43\scriptsize($\downarrow$4.8) & 67.87\scriptsize($\downarrow$4.2) &79.13\scriptsize($\downarrow$5.9) \\
			
			& & $L_2$ & 63.51\scriptsize($\downarrow$7.7) & 62.82\scriptsize($\downarrow$8.7) & 73.37\scriptsize($\downarrow$14.0) & 79.15\scriptsize($\downarrow$18.1) & 75.38\scriptsize($\downarrow$15.5) & 81.19\scriptsize($\downarrow$14.7) & 91.56\scriptsize($\downarrow$6.7) & 66.38\scriptsize($\downarrow$6.3) & 74.17\scriptsize($\downarrow$11.8) \\
			
			& & $L_3$  &59.59\scriptsize($\downarrow$13.4)  & 59.11\scriptsize($\downarrow$14.0) & 69.00\scriptsize($\downarrow$19.1) &76.95\scriptsize($\downarrow$20.4)  &73.27\scriptsize($\downarrow$17.8)  
			&77.65\scriptsize($\downarrow$18.4) &89.54\scriptsize($\downarrow$8.7) 
			&61.91 \scriptsize($\downarrow$12.6) &70.88\scriptsize($\downarrow$15.7) \\
			\midrule

			\multirow{4}{*}{\textbf{\shortstack{BD-Merging\\(Task-wise)}}} 
			& Clean & - & 70.20 & 69.44 & 83.00 & 95.35 & 89.19 & 90.70 & 97.31 & 66.17 & 82.67\\
			
			& \multirow{3}{*}{Corrupted} 
			& $L_1$ 
			&67.97\scriptsize($\downarrow$3.2)& \textcolor{blue}{\underline{67.28\scriptsize($\downarrow$3.1)}} & 80.70\scriptsize($\downarrow$2.8) & \textcolor{blue}{\underline{87.45\scriptsize($\downarrow$8.3)}} & \textcolor{blue}{\underline{81.99\scriptsize($\downarrow$8.1)}} & 84.65\scriptsize($\downarrow$6.7) &   \textcolor{blue}{\underline{93.36\scriptsize($\downarrow$4.1)}} & {\textbf{65.96\scriptsize($\downarrow$0.3)}} & \textcolor{blue}{\underline{78.67\scriptsize($\downarrow$4.8)}}\\
			
			& & $L_2$ & 65.08\scriptsize($\downarrow$7.3)  &\textcolor{blue}{\underline{64.06\scriptsize($\downarrow$7.7)}}  &\textcolor{blue}{\underline{74.00\scriptsize($\downarrow$10.8)}}  &\textcolor{blue}{\underline{80.40\scriptsize($\downarrow$15.7)}} & \textcolor{blue}{\underline{76.26\scriptsize($\downarrow$14.5)}}  & \textcolor{blue}{\underline{78.24\scriptsize($\downarrow$13.7)}}  & \textcolor{blue}{\underline{91.26\scriptsize($\downarrow$6.2)}}  & 62.13\scriptsize($\downarrow$6.1)  & \textcolor{blue}{\underline{73.93\scriptsize($\downarrow$10.6)}}  \\
			
			& & $L_3$  &{61.13\scriptsize($\downarrow$12.9)}  & \textcolor{blue}{\underline{60.63\scriptsize($\downarrow$12.7)}} &\textcolor{blue}{\underline{68.50\scriptsize($\downarrow$17.5)}} & \textcolor{blue}{\underline{78.80\scriptsize($\downarrow$17.4)}} & \textcolor{blue}{\underline{74.57\scriptsize($\downarrow$16.4)}} & \textcolor{blue}{\underline{74.65\scriptsize($\downarrow$17.7)}} &\textcolor{blue}{\underline{89.48\scriptsize($\downarrow$8.0)}} &\textcolor{blue}{\underline{58.29\scriptsize($\downarrow$11.9)}}  &\textcolor{blue}{\underline{70.76\scriptsize($\downarrow$14.4)}} \\
			\midrule
			
			\multirow{4}{*}{\textbf{\shortstack{BD-Merging\\(Layer-wise)}}} 
			& Clean & - &  73.36 & 75.24 & 89.17 & 97.75 & 92.68 & 94.90 & 98.59 & 75.53 & 87.15 \\
			
			& \multirow{3}{*}{Corrupted} 
			& $L_1$ & 71.20\scriptsize($\downarrow$2.9) 
			& {\textbf{73.03\scriptsize($\downarrow$2.9)}} 
			& {\textbf{87.17\scriptsize($\downarrow$2.2)}} 
			& {\textbf{90.05\scriptsize($\downarrow$7.9)}} &{\textbf{85.83\scriptsize($\downarrow$7.4)}}  
			&{\textbf{89.15\scriptsize($\downarrow$6.1)}}  &{\textbf{95.16\scriptsize($\downarrow$3.5)}} 
			& 74.89\scriptsize($\downarrow$0.8) 
			&{\textbf{83.31\scriptsize($\downarrow$4.4)}}\\
			
			& & $L_2$ & 
			68.43\scriptsize($\downarrow$6.7)& {\textbf{69.85\scriptsize($\downarrow$7.2)}}& {\textbf{81.00\scriptsize($\downarrow$9.2)}}& {\textbf{83.00\scriptsize($\downarrow$15.1)}}& {\textbf{80.76\scriptsize($\downarrow$12.9)}}& {\textbf{82.66\scriptsize($\downarrow$12.9)}}& {\textbf{92.65\scriptsize($\downarrow$6.0)}}& {\textbf{71.91\scriptsize($\downarrow$4.8)}}& {\textbf{78.78\scriptsize($\downarrow$9.6)}} \\
			
			& & $L_3$  &{\textbf{65.30\scriptsize($\downarrow$11.0)}} &{\textbf{66.52\scriptsize($\downarrow$11.6)}}
			&{\textbf{74.53\scriptsize($\downarrow$16.4)}}
			&{\textbf{81.30\scriptsize($\downarrow$16.8)}} &{\textbf{78.44\scriptsize($\downarrow$15.4)}} &{\textbf{78.93\scriptsize($\downarrow$16.8)}} &{\textbf{90.43\scriptsize($\downarrow$8.3)}} &{\textbf{67.45\scriptsize($\downarrow$10.7)}}
			&{\textbf{75.36\scriptsize($\downarrow$13.5)}}\\
			\midrule
	\end{tabular}}
	\label{tab:merging_bias}
\end{table*}

For each anchor \( x_i \), let \( z_i \) denote the output of the merged model \( f_{\theta^*} \). we partition its neighborhood \( \mathcal{N}_r(x_i) \) using a threshold \( \epsilon \) into a positive set
$ \mathcal{M}_r^{+}(i) = \{ j \in \mathcal{N}_r(x_i) \mid d_{ik} < \epsilon \} $ and a negative set $\mathcal{M}_r^{-}(i) = \{ j \in \mathcal{N}_r(x_i) \mid d_{ik} \geq \epsilon \}$. We define the partition function as:
\begin{equation}
	Z_i = \sum_{j \in \mathcal{M}_k^{+}(i)} \exp\left(\frac{z_i^\top z_j}{\tau}\right) + \sum_{j \in \mathcal{M}_k^{-}(i)} \exp\left(\frac{z_i^\top z_j}{\tau}\right),
\end{equation}
where \( \tau \) is a temperature hyperparameter.

The discrepancy-aware contrastive loss is computed as:
\begin{equation}
		\resizebox{\columnwidth}{!}{$
	\mathcal{L}_{Dis}^i =
	\begin{cases}
		- \log \dfrac{
			\sum\limits_{j \in \mathcal{M}_r^{+}(i)} \exp\left(\frac{z_i^\top z_j}{\tau}\right)
		}{Z_i}, & \text{if } \mathcal{M}_r^{+}(i) \neq \emptyset, \\
		- \log \left( 1 - \dfrac{
			\sum\limits_{j \in \mathcal{M}_r^{-}(i)} \exp\left(\frac{z_i^\top z_j}{\tau}\right)
		}{Z_i} \right), & \text{otherwise}.
	\end{cases}$}
\end{equation}

The final discrepancy-aware  contrastive loss is given by:
\begin{equation}
	\mathcal{L}_{Dis} = \sum_{i=1}^{N} \mathcal{L}_{Dis}^i.
\end{equation}

\paragraph{Overall Loss Function.}
To strengthen the effectiveness of merged predictions, we introduce an unsupervised objective, defined as:
\begin{equation}
	\mathcal{L}_{\mathrm{Unsup}} =  \sum_{i=1}^{N}\mathcal{H}\bigl( f_{\theta^*}(x_i)\bigr),
\end{equation}
where \(\mathcal{H}(\cdot)\) denotes the Shannon entropy.

To further enhance semantic alignment while reducing the impact of distributional shifts, we combine the unsupervised objective with the discrepancy-aware contrastive loss. Together, these losses guide the router in learning dynamic MM weights under distributionally shifted test conditions:
\begin{equation}
	\mathcal{L}_{\mathrm{BD}} = \mathcal{L}_{\mathrm{Unsup}} + \eta \mathcal{L}_{Dis},
\end{equation}
where \( \eta \) controls the strength of the  regularization.

Algorithm~\ref{alg:BDMerging} outlines the training pipeline of the proposed BDMerging framework.

\section{Experiments}

\subsection{Experimental Setting}

\paragraph{Datasets and Tasks.} Following prior works~\cite{yang2023adamerging}, we conduct experiments on eight image classification datasets: SUN397~\cite{xiao2016sun}, Cars~\cite{krause20133d}, RESISC45~\cite{cheng2017remote}, EuroSAT~\cite{helber2019eurosat}, SVHN~\cite{netzer2011reading},  GTSRB~\cite{stallkamp2011german}, MNIST~\cite{lecun2010mnist}, and DTD~\cite{cimpoi2014describing}. For each dataset \( D_k^{\mathrm{ft}} \), we construct a corresponding task by fine-tuning the CLIP model~\cite{radford2021learning} with one of the ViT backbones (ViT-B/32, ViT-B/16, or ViT-L/14). For each task~$T_k$, the merging weight~$w_k$ is learned from~$D_k^{\mathrm{aux}}$ and evaluated on~$D_k^{\mathrm{te}}$. The sets $D_k^{\mathrm{ft}}$, $D_k^{\mathrm{aux}}$, and $D_k^{\mathrm{te}}$ are non-overlapping to ensure independence from fine-tuning data.

\newcommand{\capalign}[1]{\parbox[c][2em][t]{\linewidth}{\centering #1}}

\begin{figure}[t]
	\centering
	
	\begin{subfigure}[b]{0.09\textwidth}
		\includegraphics[width=\linewidth]{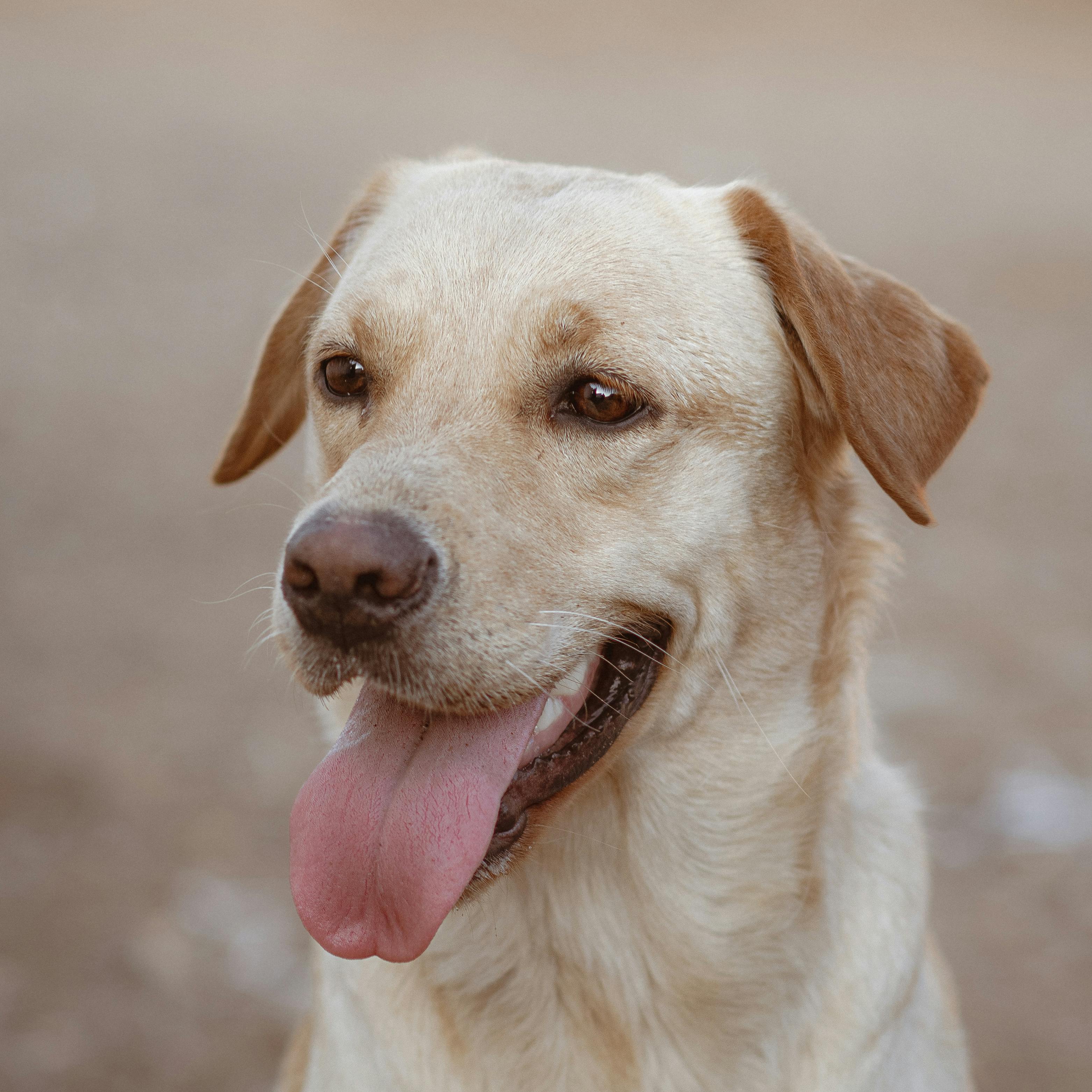}
		\caption*{\capalign{Clean}}
	\end{subfigure}
	\hfill
	\begin{subfigure}[b]{0.09\textwidth}
		\includegraphics[width=\linewidth]{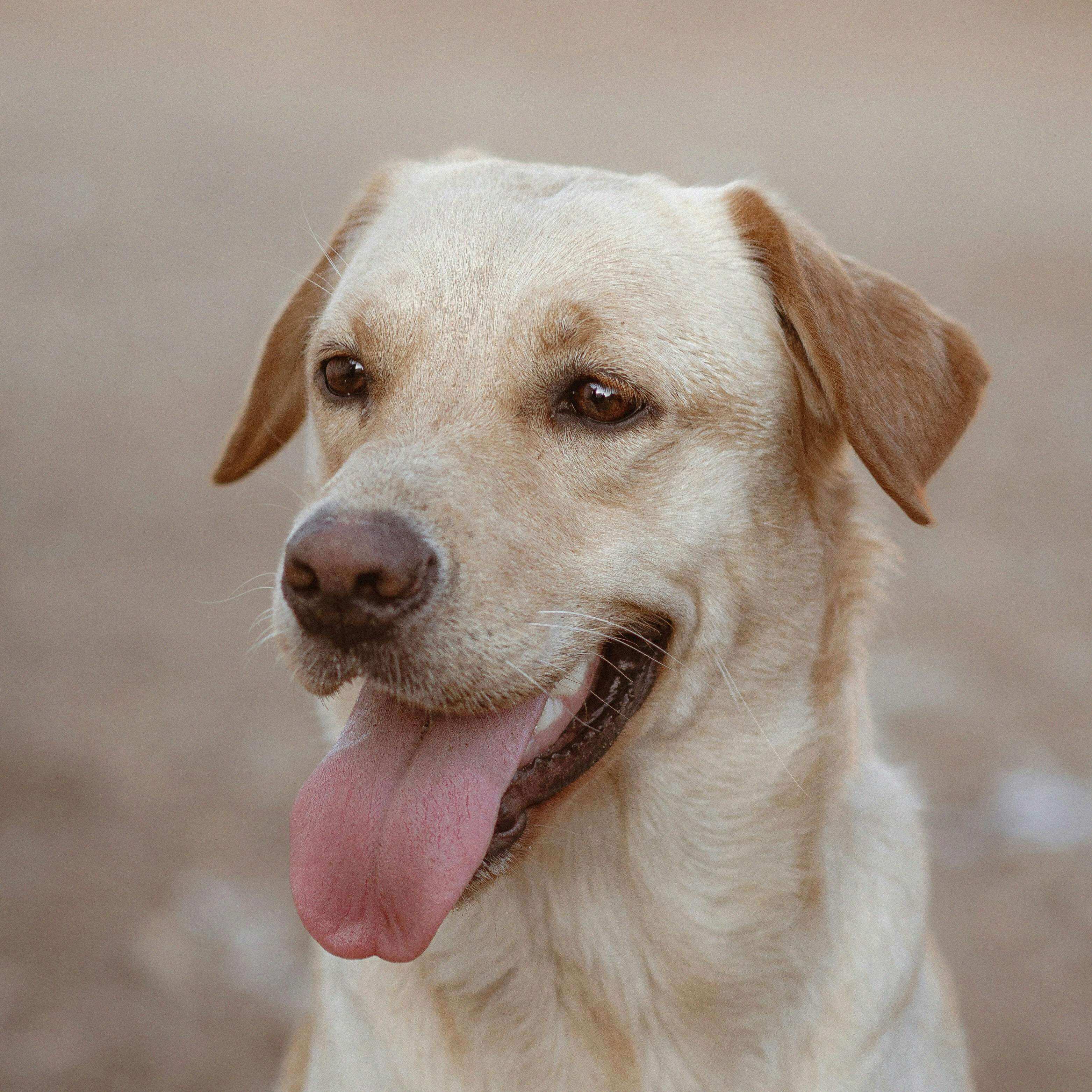}
		\caption*{\capalign{Gaussian Noise}}
	\end{subfigure}
	\hfill
	\begin{subfigure}[b]{0.09\textwidth}
		\includegraphics[width=\linewidth]{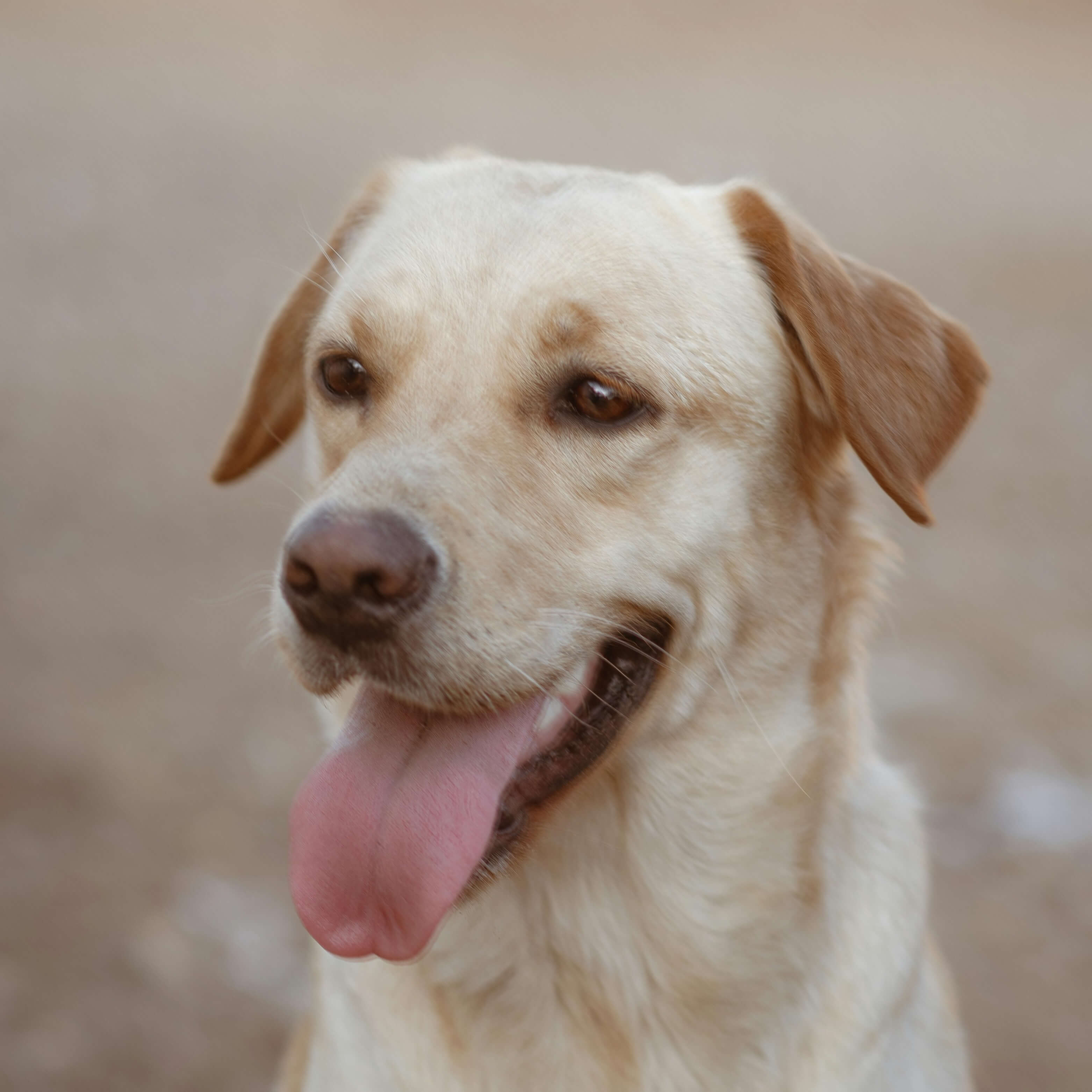}
		\caption*{\capalign{Motion Blur}}
	\end{subfigure}
	\hfill
	\begin{subfigure}[b]{0.09\textwidth}
		\includegraphics[width=\linewidth]{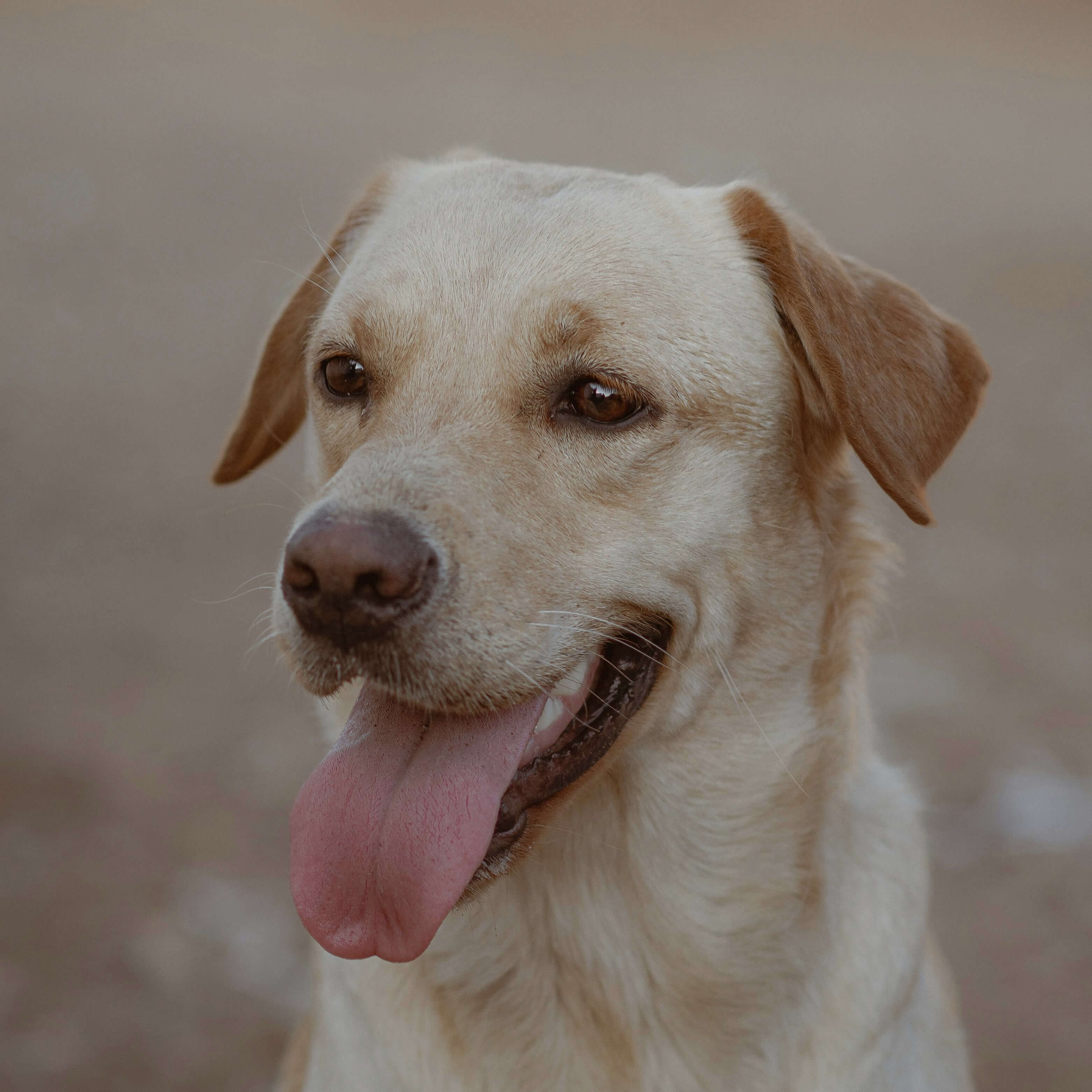}
		\caption*{\capalign{Brightness}}
	\end{subfigure}
	\hfill
	\begin{subfigure}[b]{0.09\textwidth}
		\includegraphics[width=\linewidth]{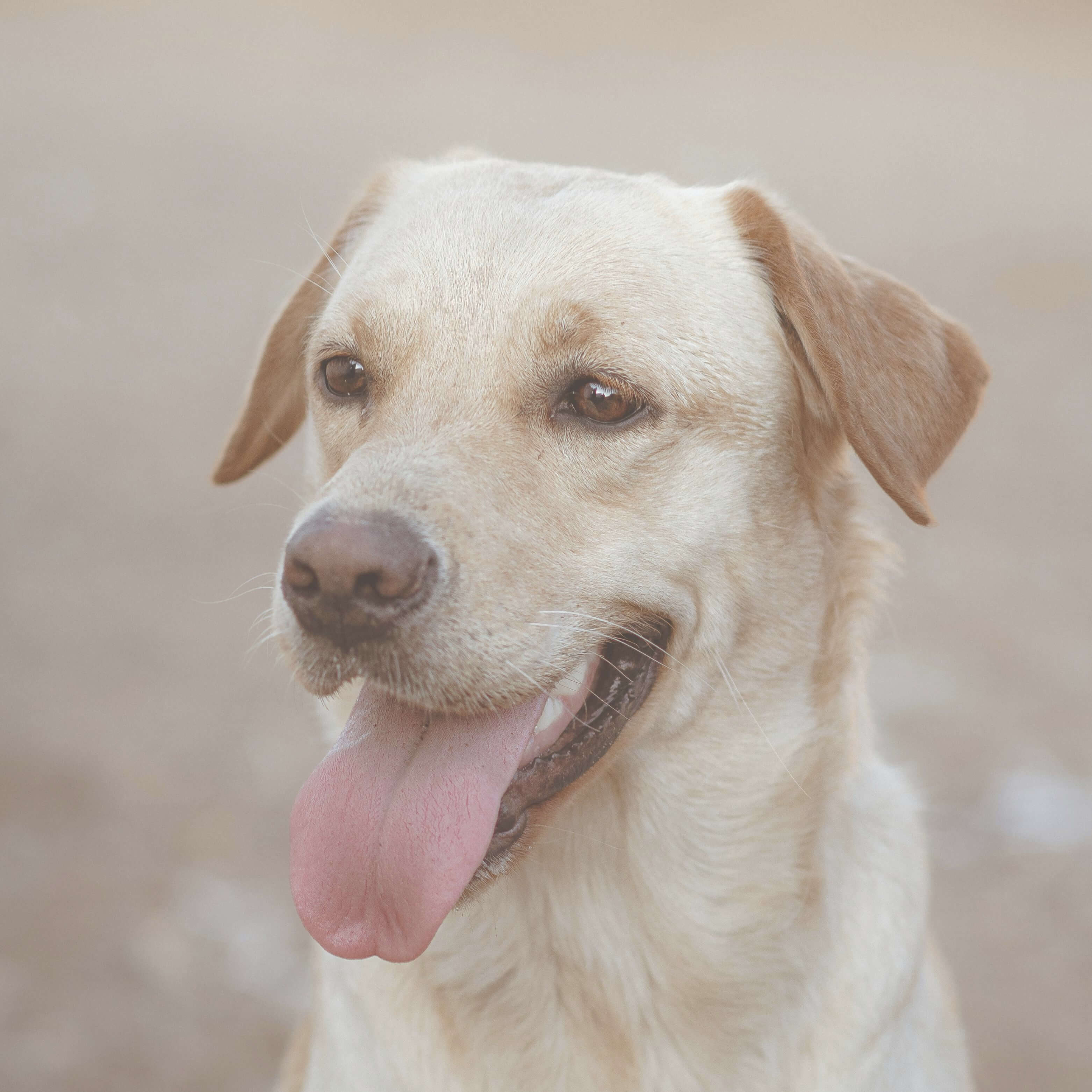}
		\caption*{\capalign{Fog}}
	\end{subfigure}
	
	\begin{subfigure}[b]{0.09\textwidth}
		\includegraphics[width=\linewidth]{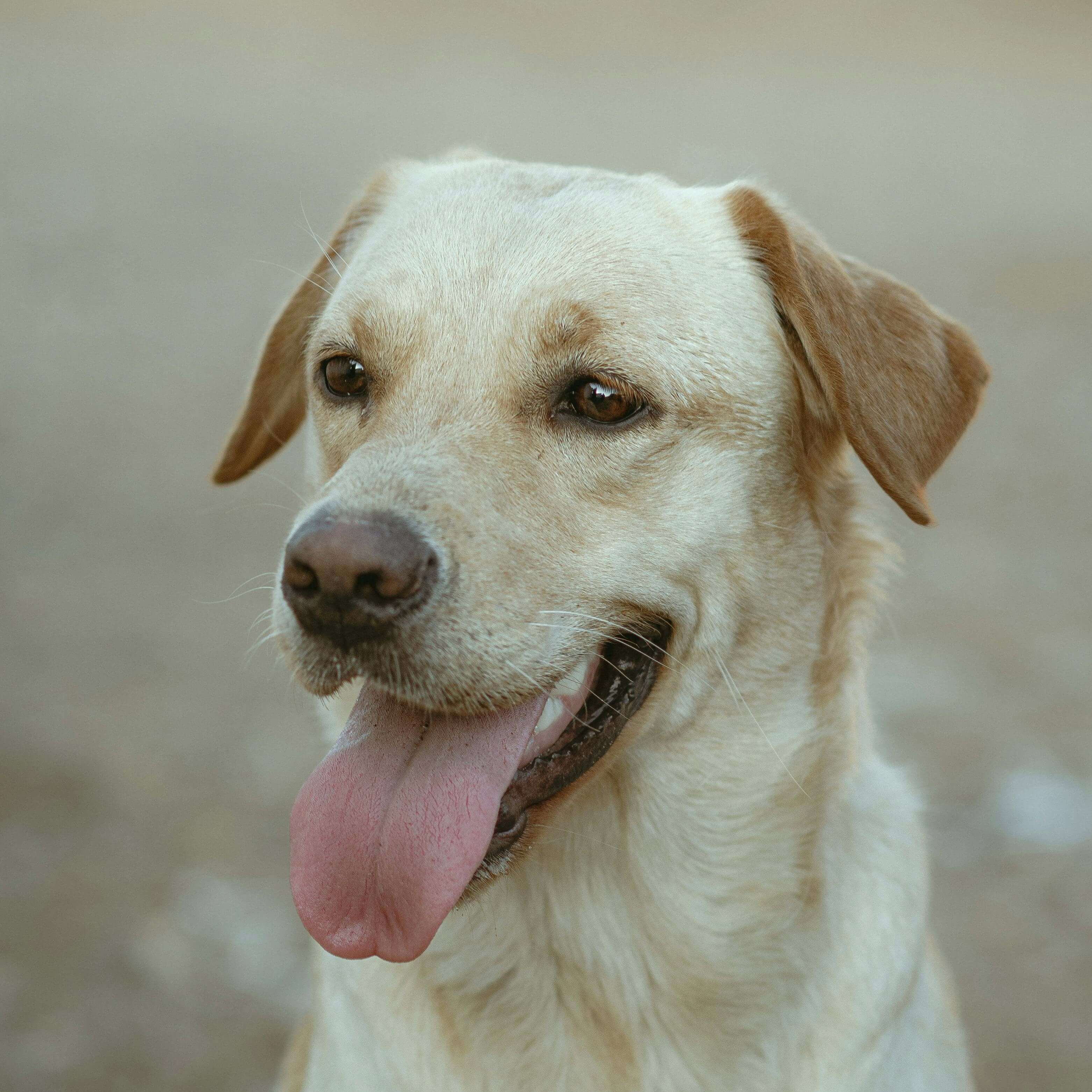}
		\caption*{\capalign{Color Shift}}
	\end{subfigure}
	\hfill
	\begin{subfigure}[b]{0.09\textwidth}
		\includegraphics[width=\linewidth]{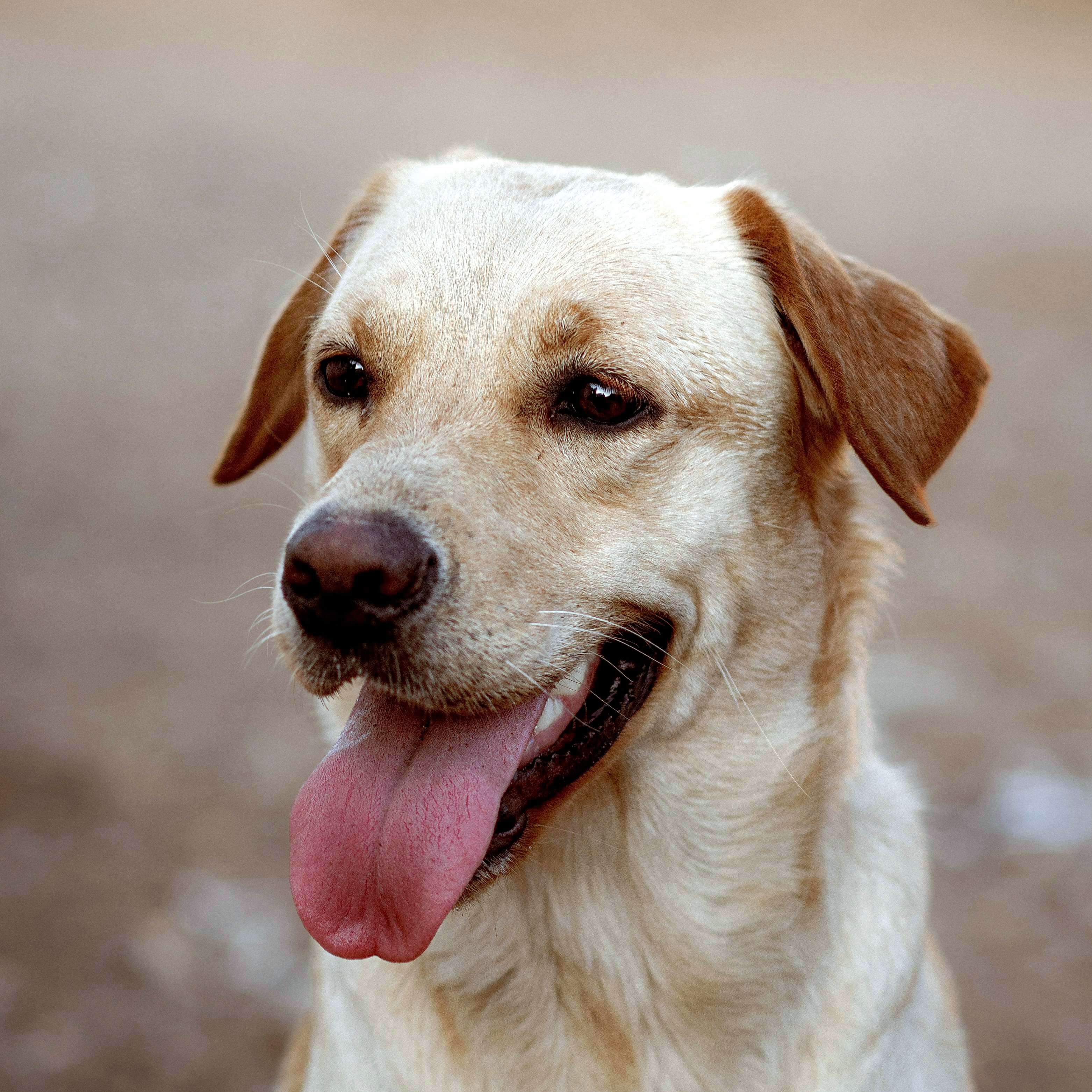}
		\caption*{\capalign{Contrast}}
	\end{subfigure}
	\hfill
	\begin{subfigure}[b]{0.09\textwidth}
		\includegraphics[width=\linewidth]{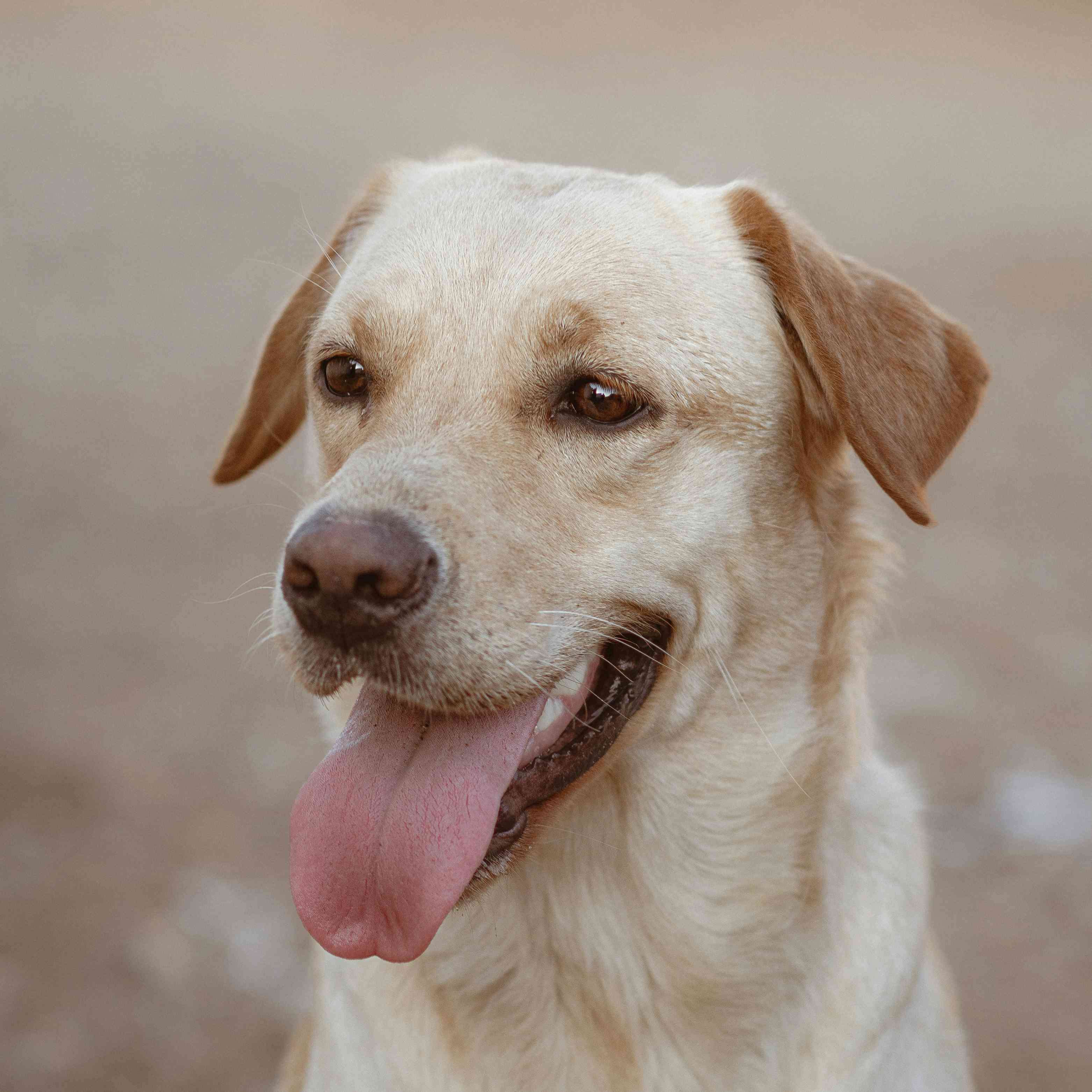}
		\caption*{\capalign{JPEG Compression}}
	\end{subfigure}
	\hfill
	\begin{subfigure}[b]{0.09\textwidth}
		\includegraphics[width=\linewidth]{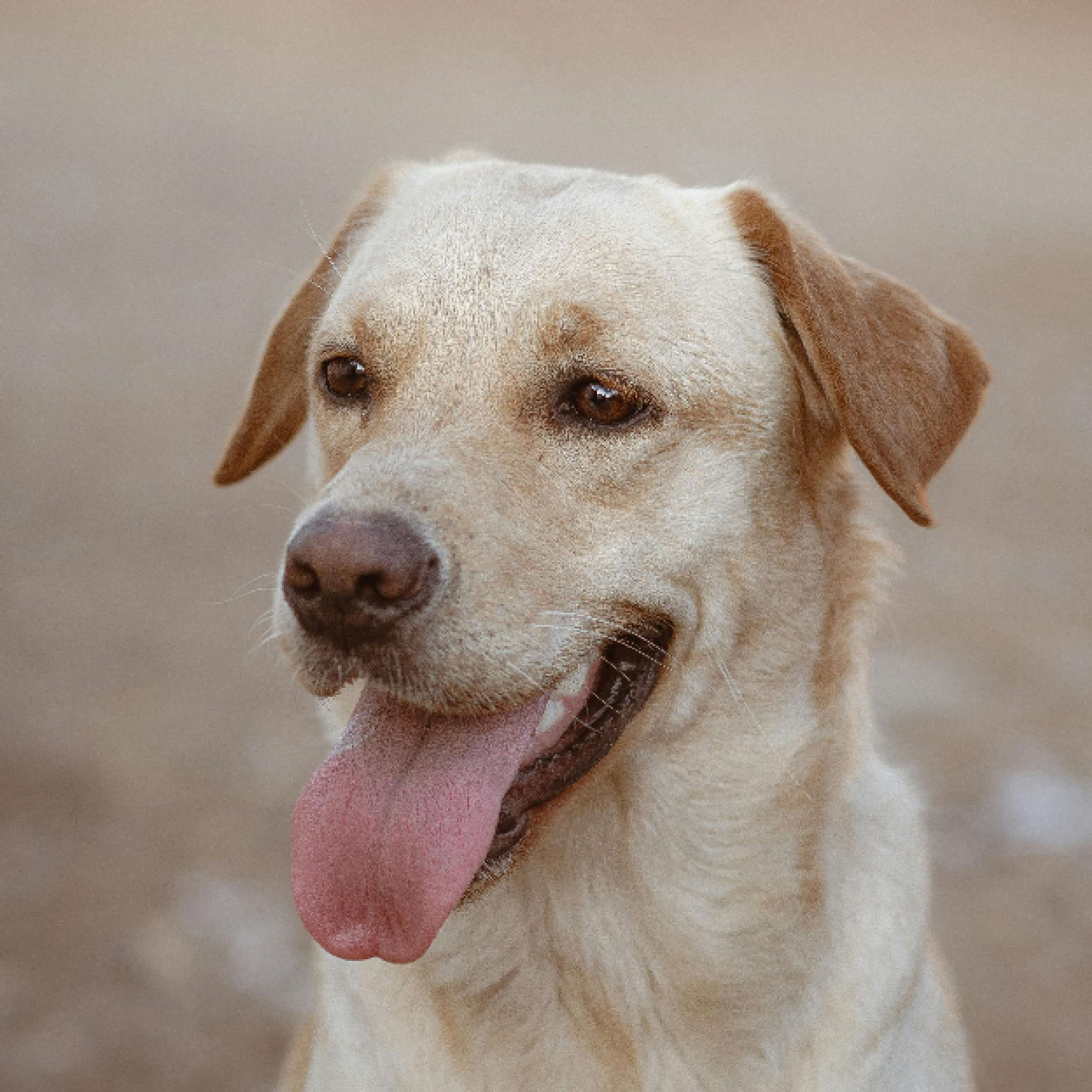}
		\caption*{\capalign{Pixelate}}
	\end{subfigure}
	\hfill
	\begin{subfigure}[b]{0.09\textwidth}
		\includegraphics[width=\linewidth]{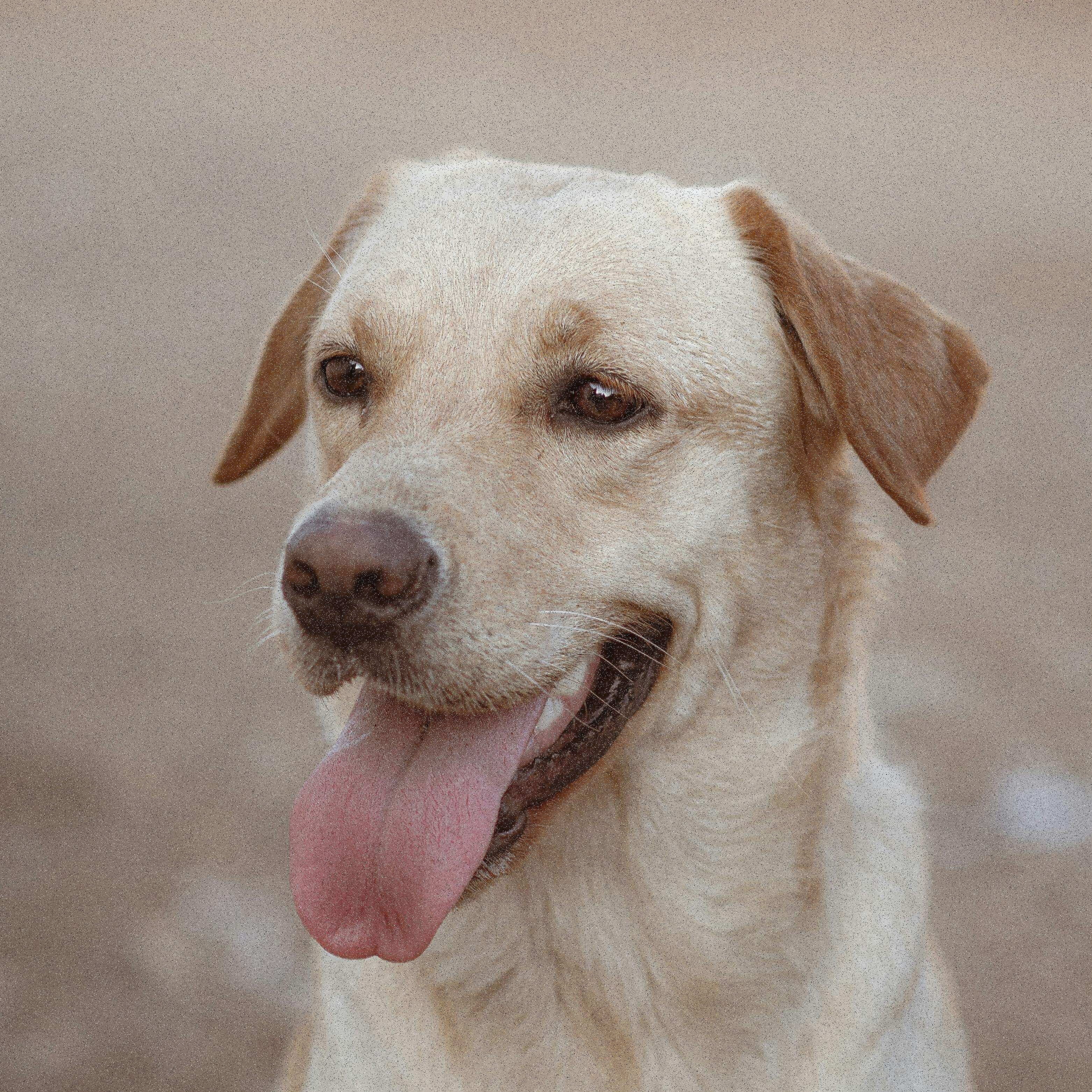}
		\caption*{\capalign{Salt\&Pepper Noise}}
	\end{subfigure}
	\vspace{-1mm}
	\caption{Visualization of corrupted image instances.}
	\label{fig:corruptions}
\end{figure}

\begin{table*}[t]
	\centering
	\small
	\caption{Performance of MM methods on seen vs. unseen tasks (4 each).}
	\vspace{-2mm}
	\resizebox{1\textwidth}{!}{%
		\begin{tabular}{l|cccc|c|cccc|c}
			\toprule
			\multirow{2}{*}{\textbf{Method}} 
			& \multicolumn{5}{c|}{\textbf{Seen Tasks}} 
			& \multicolumn{5}{c}{\textbf{Unseen Tasks}} \\
			\cmidrule(lr){2-6} \cmidrule(lr){7-11}
			& \textbf{EuroSAT \cite{helber2019eurosat}} & \textbf{SVHN \cite{netzer2011reading}} & \textbf{GTSRB \cite{stallkamp2011german}} & \textbf{MNIST \cite{lecun2010mnist}} & \textbf{Avg Acc.} & \textbf{SUN397 \cite{xiao2016sun}} & \textbf{Cars \cite{krause20133d}} & \textbf{RESISC45 \cite{cheng2017remote}}  
			&  \textbf{DTD \cite{cimpoi2014describing}} & \textbf{Avg Acc.} \\
			\midrule
			Pretrained \cite{radford2021learning}  & 43.60\scriptsize $\pm$ 0.12  &32.62\scriptsize $\pm$  0.40&32.60\scriptsize $\pm$ 0.67 &48.26 \scriptsize $\pm$ 0.56 & 39.27 \scriptsize $\pm$ 0.44&  62.46 \scriptsize $\pm$ 0.29 &59.73  \scriptsize $\pm$ 0.30& 59.60 \scriptsize $\pm$ 0.21& 46.17 \scriptsize $\pm$ 0.23& 56.99 \scriptsize $\pm$ 0.26\\
			\midrule
			Task Arithmetic \cite{ilharco2022editing}& 75.80 \scriptsize $\pm$ 0.28& 86.69 \scriptsize $\pm$ 0.17& 78.77 \scriptsize $\pm$ 0.22&97.80 \scriptsize $\pm$ 0.50&84.77 \scriptsize $\pm$ 0.29 &56.41 \scriptsize $\pm$ 0.22&52.10 \scriptsize $\pm$ 0.22&50.60 \scriptsize $\pm$ 0.49&39.15 \scriptsize $\pm$ 0.44 &49.57 \scriptsize $\pm$ 0.44\\
			Ties-Merging  \cite{yadav2023ties}   &70.85 \scriptsize $\pm$ 0.24&82.47 \scriptsize $\pm$ 0.14&70.43 \scriptsize $\pm$ 0.48&96.14 \scriptsize $\pm$ 0.21&79.97 \scriptsize $\pm$ 0.27&\textcolor{blue}{\underline{60.19 \scriptsize $\pm$ 0.42}} &\textcolor{blue}{\underline{57.07 \scriptsize $\pm$ 0.18}}&\textcolor{blue}{\underline{58.10 \scriptsize $\pm$ 0.33}}&\textcolor{blue}{\underline{42.77 \scriptsize $\pm$ 0.36}}&\textcolor{blue}{\underline{54.42 \scriptsize $\pm$ 0.32}}\\
			AdaMerging \cite{yang2023adamerging}  &89.85 \scriptsize $\pm$ 0.50 &88.14 \scriptsize $\pm$ 0.48&87.31 \scriptsize $\pm$ 0.45&97.94 \scriptsize $\pm$ 0.25&90.79 \scriptsize $\pm$ 0.32&56.36 \scriptsize $\pm$ 0.11&53.46 \scriptsize $\pm$ 0.10&51.40 \scriptsize $\pm$ 0.17&38.09 \scriptsize $\pm$ 0.35 &49.83 \scriptsize $\pm$ 0.18\\
			Twin-Merging   \cite{lu2024twin}    & \textcolor{blue}{\underline{94.83 \scriptsize $\pm$ 0.31}}& 87.10 \scriptsize $\pm$ 0.23 & \textcolor{blue}{\underline{91.60 \scriptsize $\pm$ 0.39}} &\textcolor{blue}{\underline{98.72 \scriptsize $\pm$ 0.44}} &\textcolor{blue}{\underline{93.06 \scriptsize $\pm$ 0.34}}& 59.72 \scriptsize $\pm$ 0.46&55.93 \scriptsize $\pm$ 0.29&55.63 \scriptsize $\pm$ 0.43& 40.85 \scriptsize $\pm$ 0.15 &53.03  \scriptsize $\pm$ 0.33\\
			\textbf{BD-Merging (Ours)} 
			& {\textbf{97.13 \scriptsize $\pm$ 0.38}} & {\textbf{89.20 \scriptsize $\pm$ 0.52}} & {\textbf{92.58 \scriptsize $\pm$ 0.24}} & {\textbf{99.17 \scriptsize $\pm$ 0.20}} & {\textbf{94.53 \scriptsize $\pm$ 0.33}} 
			& {\textbf{60.09 \scriptsize $\pm$ 0.26}} &{\textbf{58.13 \scriptsize $\pm$ 0.22}} & {\textbf{58.63 \scriptsize $\pm$ 0.43}} & {\textbf{43.19 \scriptsize $\pm$ 0.13}} & {\textbf{55.01 \scriptsize $\pm$ 0.26}} \\			
			\bottomrule 
	\end{tabular}}
	\label{tab:unseen}
\end{table*}
\begin{figure*}[t]
	\centering
	\includegraphics[width=0.85\linewidth]{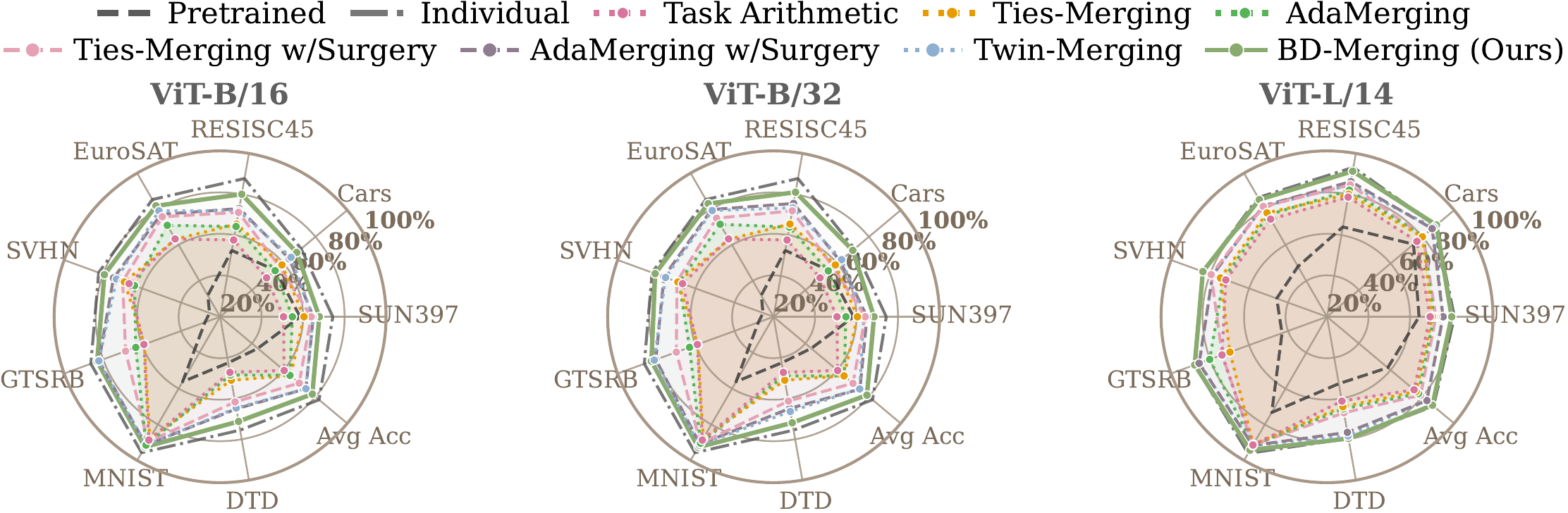}
	\caption{Evaluation of BD-Merging under test-time bias across various models.}
	\label{fig:difmodel}
\end{figure*}

\paragraph{Simulation of Realistic Test-Time Bias.}
To better approximate real-world conditions, we introduce controlled natural corruptions to the test set~$D_k^{\mathrm{te}}$, drawing on insights from prior benchmarks~\cite{hendrycks2019benchmarking}.
We group nine corruption types into three categories: sensor-induced (Gaussian noise, salt-and-pepper noise, brightness shift, and color shift), environment-induced (motion blur, fog, and contrast shift), and storage/transmission-induced (JPEG compression and pixelation). Figure~\ref{fig:corruptions} illustrates example visualizations of these corruptions.

For each task, 20\% of the test samples in~$D_k^{\mathrm{te}}$ are perturbed by randomly applying one or more corruptions from this set. To control severity, we define three levels: $L_1$ applies a single randomly selected corruption; $L_2$ applies a random combination of 1 to 5 corruptions; and $L_3$ applies a combination of 1 to 8 corruptions per sample.

\paragraph{Baselines and Evaluation Setup.}

We compare BD-Merging against several state-of-the-art MM methods, including Task Arithmetic~\cite{ilharco2022editing}, Ties-Merging~\cite{yadav2023ties}, task-wise and layer-wise AdaMerging~\cite{yang2023adamerging}, Twin-Merging~\cite{lu2024twin}, and Surgery~\cite{yang2024representation}. Unless otherwise specified, performance is reported as the average accuracy (Avg Acc) across all task-specific test sets. All methods are trained for 300 epochs with a batch size of 16, using ViT-B/32 as the shared pretrained backbone. Layer-wise weights are used by default, and corruption severity is set to $L_2$. In BD-Merging, the debiased router is implemented as a two-layer multilayer perceptron, and all regularization coefficients are set to 0.1. 

All experiments were conducted on a single NVIDIA A100 GPU with 40~GB of memory. Each experiment was conducted five times, and the results are reported as mean and standard deviation where applicable.


\subsection{Main Results}

\paragraph{Test-Time Bias.}
The experimental results of merging ViT-B/32 on eight tasks are presented in Table \ref{tab:merging_bias} under three levels of corruption severity using the realistic biased dataset. We observe that existing MM methods consistently struggle with test-time bias: even mild, commonly encountered corruptions cause noticeable drops in performance. Furthermore, all methods exhibit a clear decline in accuracy as corruption severity increases.

The BD-Merging method benefits from a reduced interference zone under perturbations, exhibiting stronger robustness compared to existing state-of-the-art merging strategies. Specifically, its task-wise and layer-wise variants show 1.8\% and 2.6\% less performance drop, respectively. We further evaluate BD-Merging across different backbone models, as shown in Figure~\ref{fig:difmodel}, where it consistently achieves the highest test-time robustness and closely approaches the performance of individual models, particularly with ViT-L/14.

\paragraph{Unseen Generalization.}
To evaluate the generalization ability of BD-Merging, we conduct experiments on a mixture of seen and unseen tasks. Figure~\ref{fig:task_comparison} shows the overall performance across different proportions of seen and unseen tasks, while Table~\ref{tab:unseen} reports detailed results of various MM methods on four specific seen and unseen tasks. While most existing state-of-the-art methods achieve strong accuracy on seen tasks, their performance drops markedly on unseen tasks, revealing limited generalization. For example, AdaMerging and Twin-Merging reach averages of 90.79\% and 93.06\% on seen tasks, respectively, but fall sharply to 49.83\% and 53.03\% on unseen tasks. This gap suggests overfitting to task-specific patterns encountered during training. 

In contrast, BD-Merging achieves a higher average accuracy of 94.53\% on seen tasks while maintaining a strong 55.01\% on unseen tasks. As shown in Figure~\ref{fig:task_comparison}, across varying ratios of seen and unseen tasks, BD-Merging consistently outperforms baselines on seen tasks and remains competitive on unseen ones. This demonstrates a superior balance between task specialization and generalization. The enhanced robustness is attributed to its debiased router, which dynamically adjusts merging weights based on features of unknown samples, effectively reducing overfitting and improving generalization under task shifts.
\begin{figure}[t]
	\centering
	\includegraphics[width=0.9\linewidth]{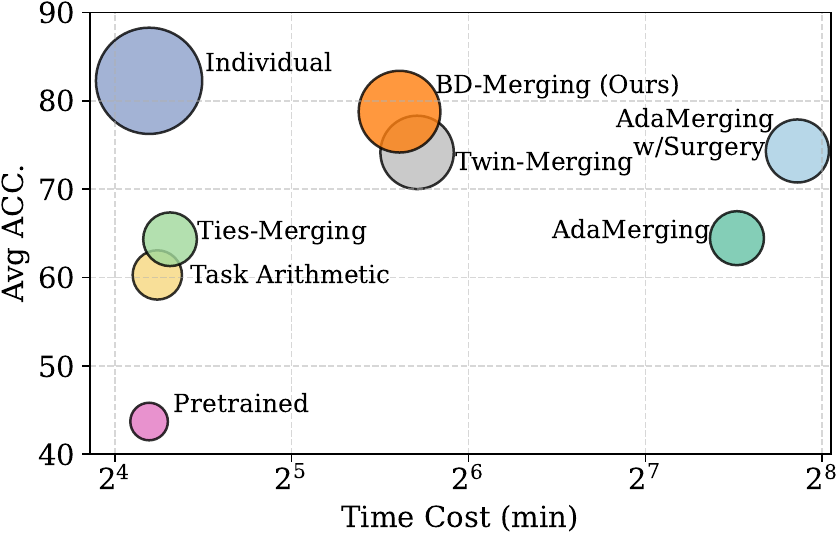}
	\vspace{-2mm}
	\caption{Performance vs. time cost of MM methods.}
	\label{fig:timecost}
\end{figure}

\paragraph{Performance vs. Time Cost.} We compare the average accuracy and time cost of various merging methods to examine their trade-offs between predictive performance and computational efficiency, as shown in Figure~\ref{fig:timecost}. 

BD-Merging achieves performance that closely approximates the individual model, while incurring substantially lower time cost. 
Although AdaMerging with Surgery attains high accuracy, it does so at the expense of significantly increased computational overhead. In contrast, Twin-Merging offers improved efficiency but suffers from reduced accuracy. These results demonstrate that BD-Merging strikes a near-optimal balance between predictive performance and computational efficiency, highlighting its practicality and scalability in real-world settings.

\subsection{Ablation Studies}

We conduct an ablation study to evaluate the contribution of each BD-Merging component under both clean and corrupted conditions, as presented in Table~\ref{tab_ablation}. 

The removal of the debiased router causes the largest performance drop, particularly under corruption, highlighting its essential role in mitigating test-time distribution shift. Excluding the ADS also causes a significant decline, with $\mathrm{Div}(\cdot)$ emerging as the most critical subcomponent, which confirms the importance of capturing diverse and reliable task-level evidence within the adjacency set. Additionally, omitting the objectives $L_{\text{Inv}}$ and $L_{\text{Dis}}$ further degrades performance, demonstrating their role in enhancing robustness. Together, each component plays a vital role in BD-Merging's overall effectiveness, especially in challenging distributionally shifted scenarios.

\begin{figure}[t]
	\centering
	\includegraphics[width=0.48\textwidth]{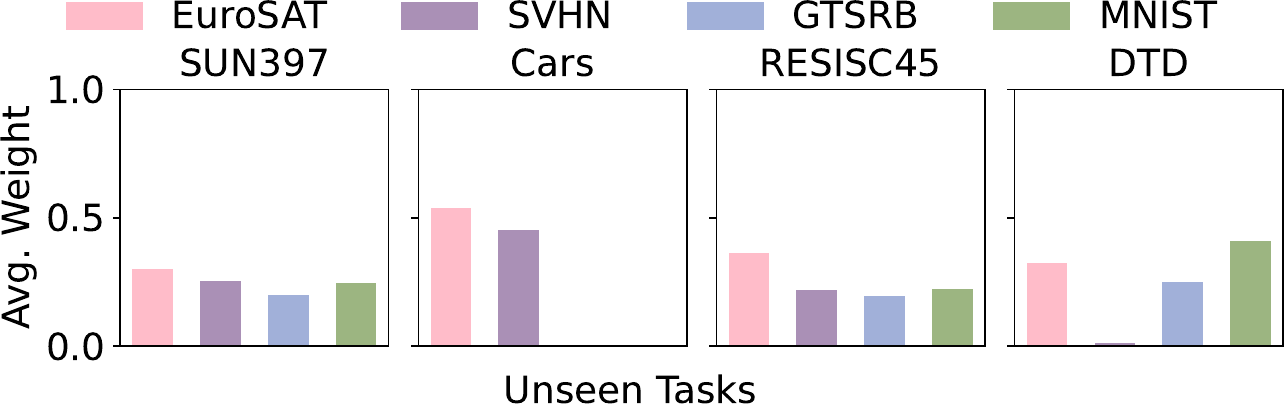}
	\vspace{-6mm}
	\caption{Variation in debiased router's average weights across unseen tasks.}

	\label{fig:router}
\end{figure}

\begin{table}[t]
	\centering
	\small
	\caption{Ablation study under clean and corrupted conditions.}
	\vspace{-1mm}
	\resizebox{0.65\columnwidth}{!}{%
		\begin{tabular}{lcc}
			\toprule		
			\textbf{Component}      & \textbf{Clean} & \textbf{Corrupted} \\ 
			\midrule
			\textbf{BD-Merging}   & \textbf{87.15\scriptsize $\pm$ 0.21} & \textbf{78.78\scriptsize $\pm$ 0.37} \\ 
			\midrule
			w/o $\mathrm{Sharp}(\cdot)$ & 86.68\scriptsize $\pm$ 0.12 & 77.94\scriptsize $\pm$ 0.15 \\ 
			w/o $\mathrm{Div}(\cdot)$   & 85.36\scriptsize $\pm$ 0.13 & 76.28\scriptsize $\pm$ 0.24 \\ 
			w/o $\mathrm{Conf}(\cdot)$  & 85.89\scriptsize $\pm$ 0.12 & 76.95\scriptsize $\pm$ 0.25 \\ 
			w/o ADS                     & 84.48\scriptsize $\pm$ 0.28 & 75.44\scriptsize $\pm$ 0.25 \\ 
			w/o Router                  & 78.31\scriptsize $\pm$ 0.33 & 67.25\scriptsize $\pm$ 0.52 \\ 
			w/o $L_{Inv}$               & 86.29\scriptsize $\pm$ 0.18 & 77.43\scriptsize $\pm$ 0.17 \\ 
			w/o $L_{Dis}$               & 83.34\scriptsize $\pm$ 0.24 & 74.26\scriptsize $\pm$ 0.36 \\ 
			\bottomrule
		\end{tabular}
	}
	\label{tab_ablation}
\end{table}

\subsection{Router Analysis}

We analyze the debiased router's behavior by visualizing the merging weights averaged over test samples from each of four unseen tasks. The results are illustrated in Figure~\ref{fig:router}.

The router produces distinct, task-specific weighting patterns that demonstrate its ability to adaptively prioritize relevant sources based on input features. These patterns offer clear interpretability by indicating the most influential sources for each task. For example, Cars shows a highly concentrated weight distribution favoring a few sources, whereas SUN397 exhibits a more balanced allocation. These findings underscore the router's capacity to capture intrinsic task characteristics and dynamically adjust merging strategies, effectively reducing task interference and enhancing robustness to distribution shifts.

\section{Conclusion}

In this paper, we introduced BD-Merging, a bias-aware unsupervised framework for model merging that addresses key challenges posed by test-time distribution shifts. BD-Merging leverages fine-grained evidence modeling to guide a debiased router, optimized via contrastive and unsupervised learning, enhancing both robustness and generalization.  This approach could improve merging efficiency while maintaining low computational overhead. Extensive experiments demonstrate that BD-Merging consistently outperforms state-of-the-art baselines across diverse distribution shift scenarios, highlighting its practical potential for real-world applications.
\section*{Acknowledgment}
This work was supported by the National Key R\&D Program of China (2023YFA1009500).

{
    \small
    \bibliographystyle{ieeenat_fullname}
    \bibliography{main}
}


\end{document}